\documentclass[10pt,journal,compsoc]{IEEEtran}

\usepackage[nocompress]{cite}

\usepackage{url}
\usepackage{multirow}

\usepackage{siunitx}
\usepackage{algorithm}
\usepackage{algorithmic}

\usepackage{pifont}
\newcommand{\cmark}{\ding{51}}%
\newcommand{\xmark}{\ding{55}}%

\usepackage[compact]{titlesec}

\newcommand{\TT}[1]{\texttt{#1}}

\usepackage{xcolor}
\usepackage{fancyhdr,graphicx}
\usepackage{amssymb,amsthm,amsmath}
\usepackage{booktabs,array} 
\usepackage{thmtools}
\usepackage{thm-restate}
\usepackage{tikz}
\usepackage{resizegather}
\usepackage{wrapfig}
\usepackage{enumitem,kantlipsum}
\usepackage{mathtools}
\usepackage{nccmath}

\newtheoremstyle{mystyle}
{6pt} 
{\topsep} 
{} 
{} 
{\bfseries} 
{.} 
{.5em} 
{} 

\theoremstyle{definition}

\theoremstyle{plain}

 \usepackage{xspace}

\usepackage{dsfont}
\usepackage{stmaryrd}

\renewcommand{\phi}{\varphi}

\newcommand{\abs}[1]{\vert {#1} \vert}
\newcommand{\norm}[1]{\Vert {#1} \Vert}

\newcommand{\normsq}[1]{\norm{#1}^2}

\newcommand{\Expect}{\operatorname{\mathbb{E}}}

\newcommand{\loss}{\mathcal{L}}

\usepackage{amsmath,amsfonts,bm}

\def\eqref#1{equation~\ref{#1}}

\def\1{\bm{1}}

\def\va{{\bm{a}}}
\def\vb{{\bm{b}}}

\DeclareMathAlphabet{\mathsfit}{\encodingdefault}{\sfdefault}{m}{sl}
\SetMathAlphabet{\mathsfit}{bold}{\encodingdefault}{\sfdefault}{bx}{n}

\def\sR{{\mathbb{R}}}

\newcommand{\E}{\mathbb{E}}

\DeclareMathOperator{\sign}{sign}

\newcommand{\bitwidth}{\mathcal{B}}
\DeclareMathOperator*{\logquant}{Q_{log}}

\DeclareMathOperator*{\SR}{\mathrm{SR}}

\newcommand{\basefactor}{\gamma}
\DeclareMathOperator*{\uquant}{Q_U}
\DeclareMathOperator*{\wquant}{Q_W}
\DeclareMathOperator*{\aquant}{Q_A}
\DeclareMathOperator*{\equant}{Q_E}
\DeclareMathOperator*{\gquant}{Q_G}

\usepackage[para,online,flushleft]{threeparttable}

\begin{document}

\title{LNS-Madam: Low-Precision Training 
\\ in Logarithmic Number System \\ using Multiplicative Weight Update}

\author{Jiawei Zhao, Steve Dai, Rangharajan Venkatesan, Brian Zimmer, Mustafa Ali, Ming-Yu Liu, Brucek Khailany, William J. Dally, Anima Anandkumar%
\IEEEcompsocitemizethanks{\IEEEcompsocthanksitem J. Zhao and A. Anandkumar are with Caltech.\\
E-mail: jiawei@caltech.edu
\IEEEcompsocthanksitem S. Dai, R. Venkatesan, B. Zimmer, M. Liu, B. Khailany, B. Dally, A. Anandkumar are with NVIDIA.
\IEEEcompsocthanksitem M. Ali is with Purdue University.}%
}

\vskip 0.3in

\IEEEtitleabstractindextext{%
\begin{abstract}

Representing deep neural networks (DNNs) in low-precision is a promising approach to enable efficient acceleration and memory reduction. 
Previous methods that train DNNs in low-precision typically keep a copy of weights in high-precision during the weight updates.
Directly training with low-precision weights leads to accuracy degradation due to complex interactions between the low-precision number systems and the learning algorithms.  
To address this issue, we develop a co-designed low-precision training framework, termed LNS-Madam, in which we jointly design a logarithmic number system (LNS) and a multiplicative weight update algorithm (Madam). 
We prove that LNS-Madam results in low quantization error during weight updates, leading to stable performance even if the precision is limited.
We further propose a hardware design of LNS-Madam that resolves practical challenges in implementing an efficient datapath for LNS computations.
Our implementation effectively reduces energy overhead incurred by LNS-to-integer conversion and partial sum accumulation.
Experimental results show that LNS-Madam achieves comparable accuracy to full-precision counterparts with only 8 bits on popular computer vision and natural language tasks.
Compared to FP32 and FP8, LNS-Madam reduces the energy consumption by over 90\% and 55\%, respectively.

\end{abstract}
}

\maketitle

\IEEEraisesectionheading{\section{Introduction}\label{sec:introduction}}
\IEEEPARstart{D}{eep} neural networks (DNNs) have shown impressive performance in many applications, including image classification and language processing. However, training and deploying DNNs typically incurs significant computation and energy costs.
Traditionally, values in neural networks are represented using floating-point (32-bit) numbers, which incurs a large arithmetic and memory footprint, and hence significant energy consumption. However, recent studies suggest that high-precision number formats are redundant, and models can be quantized in low-precision with little loss in accuracy \cite{fp16,8bit}. Low-precision numbers only require low-bitwidth computational units, leading to better computational efficiency and less required memory bandwidth and capacity.

While low-precision training methods generally reduce computational costs, energy efficiency can be further improved by choosing a logarithmic number system (LNS) for representing numbers. 
LNS achieves a higher computational efficiency by transforming expensive multiplication operations in the network layers to inexpensive additions in their logarithmic representations.
In addition, it attains a wide dynamic range and can provide a good approximation of the non-uniform weight distribution in neural networks. Thus LNS is an excellent candidate for training DNNs in low-precision \cite{frankle2018lottery,nanoconnectome,conv_log}.

Although previous studies demonstrate that it is feasible to train networks in low-precision using LNS, these approaches have not yet shown promising results on larger datasets and state-of-the-art models \cite{conv_log,4_bit_ultra_low}. 
Standard LNS fixes the base of the logarithm, termed log-base, to be precisely two. However, a more flexible log-base is needed since the numbers in DNNs require different quantization gaps during training \cite{otherlog}.
A flexible log-base can introduce additional hardware overhead due to expensive conversion operations between the logarithmic and integer (linear) formats.
This motivates us to design a LNS that has a flexible choice of the log-base while maximizing the efficiency of LNS-to-integer conversions.


Conventional low-precision training methods typically require high-precision copies of weights and gradients during weight update to maintain optimization stability. 
In fact, most recent studies even use a full-precision (FP32) copy of weights \cite{conv_log,4_bit_ultra_low}. 
This introduces additional energy costs and expensive FP32 arithmetic, which becomes prohibitive especially in energy-constrained edge devices.

\begin{figure}[t!]
    \centering
    \includegraphics[width=0.8\linewidth]{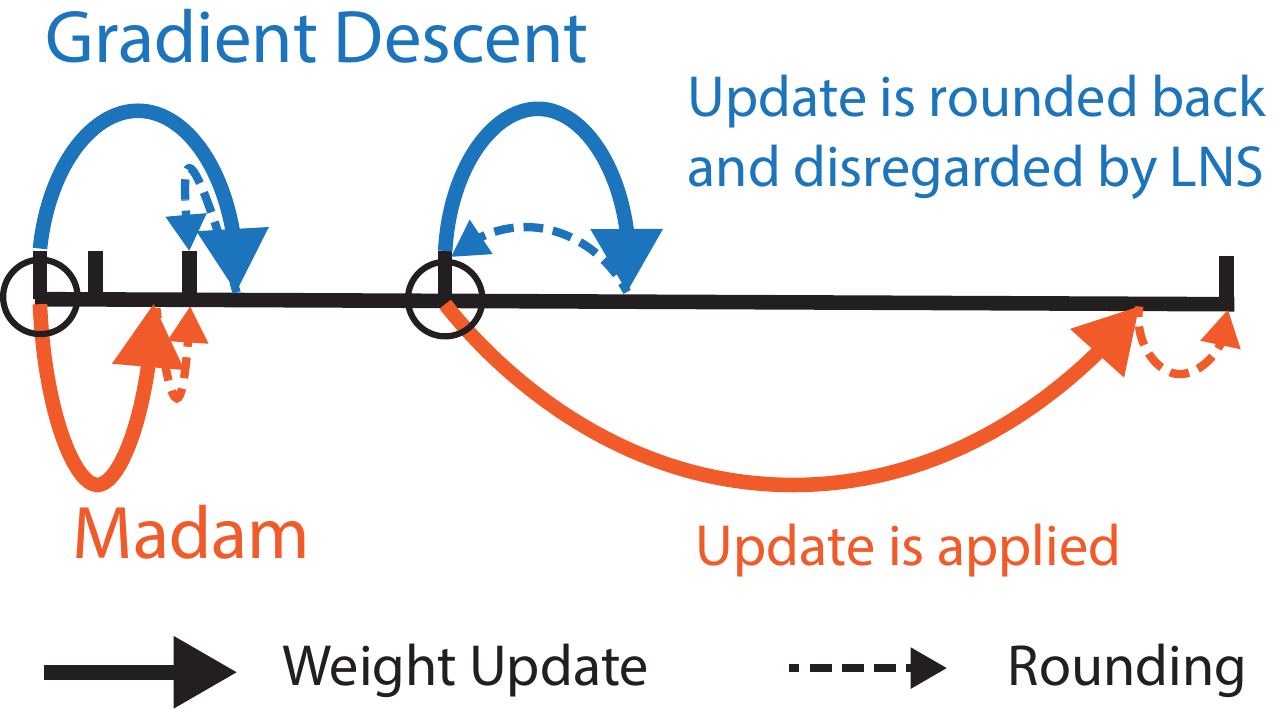}
    \caption{Illustration for updating weights using Gradient Descent (GD) and Madam under logarithmic representation. Each coordinate represents a number stored in LNS. Assume the weights at two circles receive the same gradient. The updates generated from GD are disregarded as the weights move larger, whereas the updates generated by Madam are adjusted with the weights.}
    \label{fig:add_vs_mul}
\end{figure}
The high-precision requirement for weight updates is due to complex interactions between learning algorithms and number systems, which has usually been ignored in previous studies. 
For example, as illustrated in Fig. \ref{fig:add_vs_mul}, updates generated by stochastic gradient descent (SGD) are disregarded more frequently by LNS when the weights become larger. 
This is because the quantization gap grows exponentially as the weights become larger in LNS, which suggests that conventional learning algorithms may not be suitable for LNS. 
Hence, in previous studies, high-precision weight copies are required to avoid numerical instabilities \cite{conv_log,otherlog}.

To directly update the weights in low-precision, we employ a learning algorithm tailored to LNS. 
Recently, Bernstein et al. \cite{bernstein2020learning} proposed the Madam optimizer based on multiplicative updates, which is equivalent to updating weights additively in the logarithmic space. 
As illustrated in Fig. \ref{fig:add_vs_mul}, Madam generates larger magnitudes of the updates when the weights become larger, making it more suitable for a logarithmic weight representation.
However, Bernstein et al. \cite{bernstein2020learning} still employ full-precision training with Madam without considering low-precision LNS. 

In this work, we propose a co-designed low-precision training framework called LNS-Madam in which we adopt LNS (with a more flexible log-base) for representing DNNs and apply a modified Madam (tailored to LNS) to train these networks. LNS-Madam reduces the precision requirements for all components of the training, including \emph{forward and backward propagation, as well as weight updates}.

\begin{figure}[t!]
    \centering
    \includegraphics[scale=0.4]{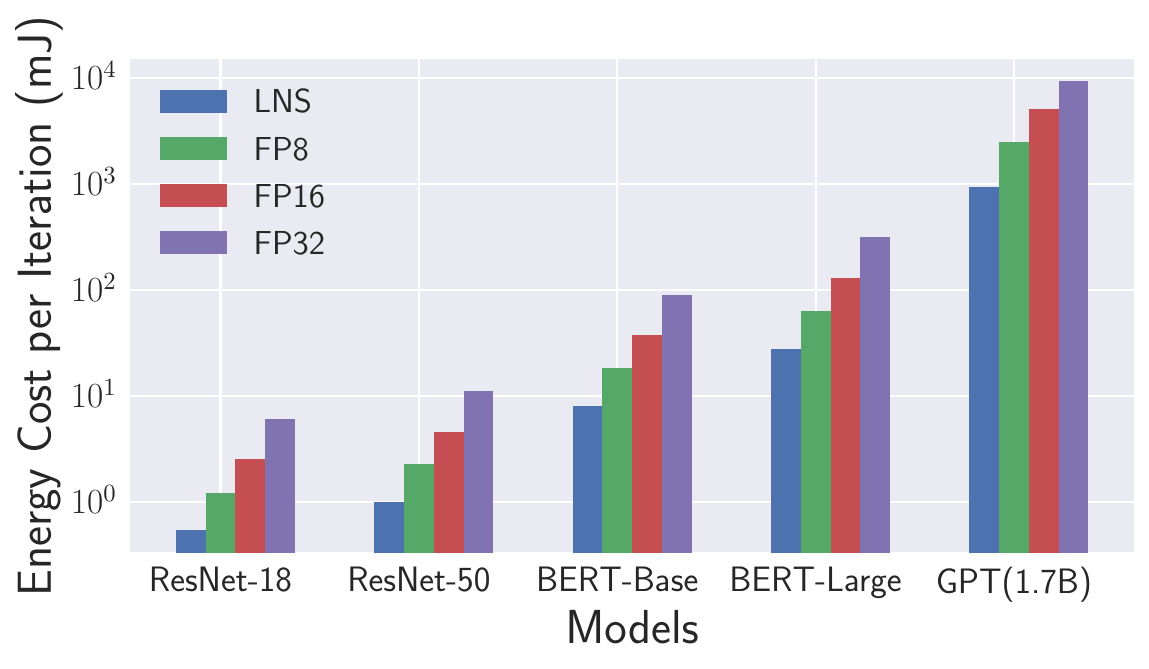}
    \vskip -0.13in
    \caption{Energy efficiency for training different models with various number formats. The per-iteration energy consumption (mJ) is listed.}
    \label{fig:energy_models}
\end{figure}

\textbf{Our contributions are summarized as follows:}
\begin{enumerate}[leftmargin=*]
    \item We design a multi-base LNS where the log-base can be fractional powers of two. The multi-base LNS accommodates the precision and range requirements of the training dynamics while being hardware-friendly. In addition, we propose an approximation for the addition arithmetic in LNS to further improve its energy efficiency.

    \item We propose an efficient hardware implementation of LNS-Madam that addresses challenges in designing an efficient datapath for LNS computations, including accumulation and conversion between logarithmic and integer formats. We leverage this implementation to study the energy benefits of training in LNS.

    \item To achieve low-precision weight updates in LNS, we replace standard SGD or Adam optimizers with our proposed optimizer based on Madam, which directly updates the weights in the LNS. Through theoretical analysis and empirical evaluations, we show that the proposed Madam optimizer achieves significantly lower quantization error in LNS. 

    \item In our experiments, LNS-Madam achieves full-precision accuracy with 8 bit representations on popular computer vision and natural language tasks while reducing energy consumption by over 90\%. The energy efficiency results for training different models with various number formats are summarized in Fig. \ref{fig:energy_models}.
    
    

\end{enumerate}




\setlength{\belowdisplayskip}{2pt}
\setlength{\belowdisplayshortskip}{2pt}
\setlength{\abovedisplayskip}{2pt}
\setlength{\abovedisplayshortskip}{2pt}

\section{Hardware-friendly Multi-Base Logarithmic Number System}
\label{sec:LNS}

In this section, we introduce our multi-base logarithmic number system (LNS), including the corresponding number representation and arithmetic operations.

We start with the mathematical formulation that we will use throughout this paper. We assume that the DNN $F(\cdot,W)$ is composed of $L$ layers with learnable weights $W$ and input activations $X$ across the layers. $\loss(W)$ denotes the training loss. The forward propagation is defined as: $X_{l} = f_{l}(X_{l-1}, W_{l})$, where $l \in[1, L]$ denotes layer $l$. $\nabla_{X_l} = \frac{\partial \loss(W)}{\partial X_l}$ and $\nabla_{W_l} = \frac{\partial \loss(W)}{\partial W_l}$ denote gradients with respect to input activations and weights, respectively, at layer $l$. For a number system, we define $\bitwidth$ as the bitwidth, $x$ as a number, and $x^{q}$ as the number in quantized format.

\subsection{Multi-base Logarithmic Representation} 
Unlike prior work that uses exactly 2 as the base of the logarithmic representation, we propose a multi-base logarithmic representation that allows the base to be two with a fractional exponent, such that:
\begin{equation*}
    x = \sign \times 2^{\Tilde{x} / \basefactor} \quad \Tilde{x} = 0,1,2,...,2^{\bitwidth - 1}-1,
\end{equation*}
where $\Tilde{x}$ is an integer and $\basefactor$ is the base factor that controls the fractional exponent of the base. $\basefactor$ controls the quantization gap, which is the distance between successive representable values within the number system. Previous work has already demonstrated that logarithmic quantized neural networks achieve better performance when relaxing $\basefactor$ from $1$ to $2$ \cite{conv_log}. 
We find that further relaxation can help adapt to different models and datasets.
Therefore we generalize the base factor setting, enabling more flexibility in controlling the quantization gap in order to more accurately approximate the training dynamics.
In addition, we specially restrict $\basefactor$ to be powers of 2 for hardware efficiency, as described later.
\subsection{Arithmetic Operations in LNS}
\label{subsec:log_operations}
One of the benefits of using LNS stems from the low computational cost of its arithmetic operations. We use dot product operations as an example since they are prevalent during training. Consider two vectors $\va \in \sR^n$ and $\vb \in \sR^n$ that are represented by their integer exponents $\Tilde{\va}$ and $\Tilde{\vb}$ in LNS. A dot product operation between them can be represented as follows:
\begin{equation}
\label{eq:dot_product}
\begin{split}
        \va^{T} \vb & = \sum_{i=1}^{n} \sign_i \times 2^{\Tilde{\va_i}  / \basefactor} \times 2^{\Tilde{\vb_i}  / \basefactor} \\
        & = \sum_{i=1}^{n} \sign_i \times 2^{(\Tilde{\va_i}+\Tilde{\vb_i})  / \basefactor} \\
        & = \sum_{i=1}^{n} \sign_i \times 2^{\Tilde{p_i} / \basefactor}, 
\end{split}
\end{equation}
where $\sign_i = sign(\va_i) \oplus sign(\vb_i)$. In this dot product operation, each element-wise multiplication is computed as an addition between integer exponents, which significantly reduces the computational cost by requiring adders instead of multipliers. 

While the multiplications are easy to compute in LNS, the accumulation is difficult to compute efficiently as it requires first converting from logarithmic to integer format and then performing the addition operation. 
The conversion between these formats is generally expensive as it requires computing $2^{\Tilde{p_i}/ \basefactor}$ using polynomial expansion.
To overcome this challenge, we decompose the exponent $2^{\Tilde{p_i} / \basefactor}$ into a quotient component $\Tilde{p_i}_q$ and and a remainder component $\Tilde{p_i}_r$ as follows:
\begin{equation}
\label{eq:decomposition}
2^{\Tilde{p_i} / \basefactor}=2^{\Tilde{p_i}_q + \Tilde{p_i}_r/ \basefactor}=2^{\Tilde{p_i}_q} \cdot 2^{\Tilde{p_i}_r/ \basefactor}.
\end{equation}
With this decomposition, converting from LNS to integer requires only a table lookup for $2^{\Tilde{p_i}_r/ \basefactor}$ followed by a shift for $2^{\Tilde{p_i}_q}$.
For the table lookup, we simply maintain $\gamma$ constants $2^{i/\gamma}\;\forall i\in\{0,1,...,\gamma-1\}$ and select the constant based on the remainder $\Tilde{p_i}_r$.
Note that because $\gamma$ is restricted to be power of 2, the remainder can be efficiently extracted from the least-significant bits (LSB) of the exponent while the quotient can be extracted from the most-significant bits (MSB).
Typically the lookup table (LUT) requires $2^{\bitwidth}$ entries for storing all possible values.

\subsection{Conversion Approximation}
In addition to the exact conversion technique discussed above, we can further reduce the cost of the LNS-to-integer conversion using a hybrid approximation method.
Our method is based on Mitchell approximation \cite{5219391}: $2^{\Tilde{x}/\gamma} \approx (1 + \Tilde{x}/\gamma)$, where the logarithmic format can be efficiently approximated to the integer format when $\Tilde{x}/\gamma$ is small.
Specifically, we further split the remainder into a LSB and a MSB component.
The value of the LSB is approximated using Mitchell approximation, and the value of the MSB is performed with table lookup.
This helps reduce the size of the LUT. 
We present a detailed description of our approximation method in the Appendix~\ref{apdx:lns}.
In addition, since the approximation serves as an additional non-linear operation in neural networks, we find the approximated training does not damage accuracy in practice. 

\setlength{\belowdisplayskip}{2pt}
\setlength{\belowdisplayshortskip}{2pt}
\setlength{\abovedisplayskip}{2pt}
\setlength{\abovedisplayshortskip}{2pt}

\section{Quantized Forward and Backward Propagation on LNS} 

\label{sec:low-precison-training}
\begin{figure*}[t!]
    \centering
    \centerline{\includegraphics[scale=0.5]{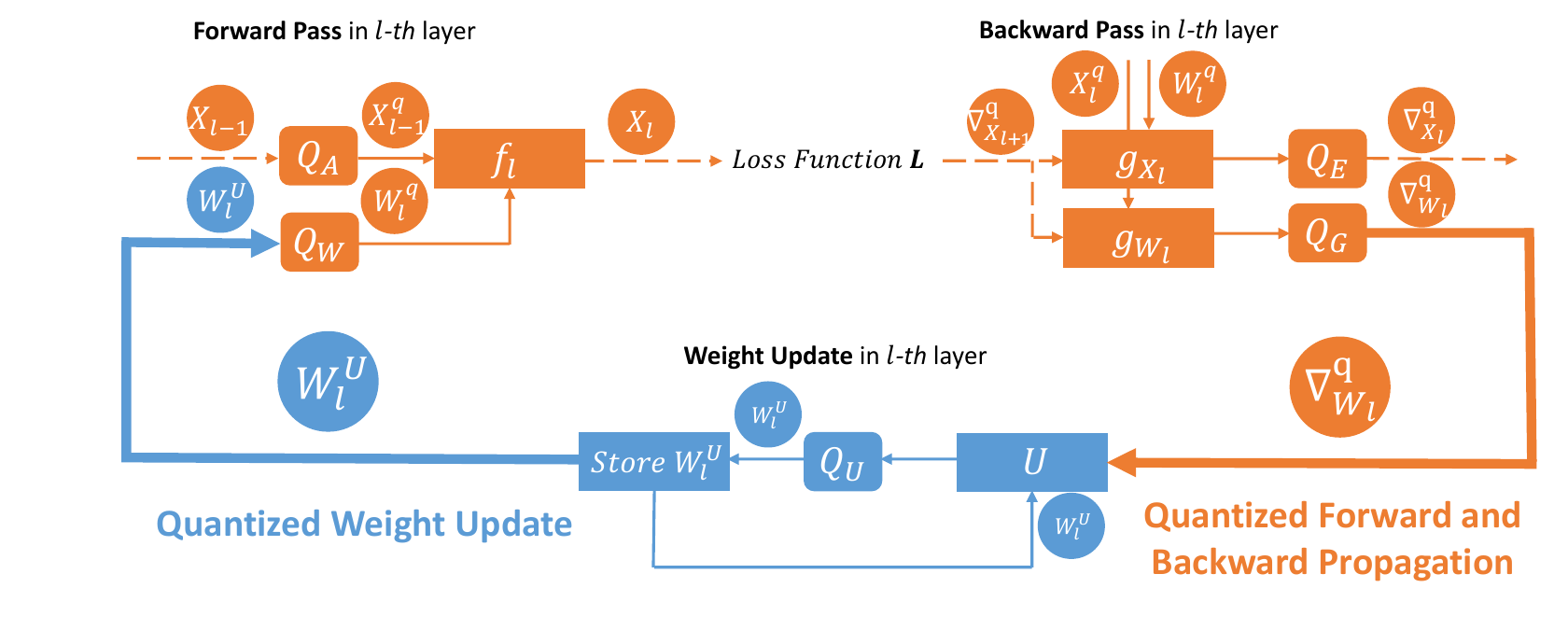}}
    \vskip -0.2in
    \caption{Illustration of LNS-Madam. Quantized training includes quantizing weights $W$ and activations $X$ in forward propagation, and weight gradients $\nabla_{W}$ and activation gradients $\nabla_{X}$ in backward propagation. $g_{X}$ and $g_{W}$ denote the functions to compute gradients. Quantized weight update applies a quantization function $Q_{U}$ over weights after any learning algorithm $U$ updates them. The quantized weights $W^{U}$ are the actual numbers stored in the system.}
    \label{fig:framework}
    \vskip -0.02in
\end{figure*}

In this section, we introduce how to apply multi-base LNS to quantized training, as illustrated in Fig. \ref{fig:framework}.



To realize reduced precision for values and arithmetic during training, we define a logarithmic quantization function $\logquant: \sR \to \sR$, which quantizes a real number into a sign and an integer exponent using a limited number of bits. $\logquant$ is defined as follows:
\begin{equation}
\label{eq:log_quant}
    \logquant(x) = sign(x) \times s \times 2^{(\Tilde{x} / \basefactor)},
\end{equation}
where $\Tilde{x} = clamp(\, round(\, \log_2 (\abs{x} / s) \times \gamma), 0, 2^{\bitwidth - 1}-1)$, and $s \in \sR$ denotes a scale factor. $\logquant$ first brings scaled numbers $\abs{x} / s$ into their logarithmic space, magnifies them by the base factor $\basefactor$ and then performs rounding and clamping functions to convert them into desired integer exponents $\Tilde{x}$. The scale factor $s$ usually is shared within a group of numbers, and its value is assigned to match the maximum number within the group. 

We apply quantization-aware training (QAT) for quantizing weights and activations during forward propagation. Each quantizer is associated with a STE to allow the gradients to directly ﬂow through the non-differentiable quantizer during backward pass \cite{bengio2013estimating}. Because QAT views each quantization function as an additional non-linear operation in the networks, the deterministic quantization error introduced by any quantizer in the forward pass is implicitly reduced through training. We define weight quantization function as $\wquant$ and activation quantization function as $\aquant$ for each layer during forward propagation, where $W_l^{q} = \wquant \left(W_l\right)$ and $X_l^{q} = \aquant \left(f_{l}\left(X_{l-1}^{q}, W_l^{q} \right)\right)$.


In order to accelerate training in addition to inference, gradients also need to be quantized into low-precision numbers. As shown by recent studies, the distribution of gradients resembles a Gaussian or Log-Normal distribution \cite{chmiel2021neural, bernstein2018signsgd}. This suggests that logarithmic representation may be more suitable than fixed-point representations when quantizing gradients to attain hardware efficiency. We quantize the activation gradients using quantization function $\equant$: $\nabla_{X_l}^{q} = \equant \left( \nabla_{X_l} \right)$. We also quantize the weight gradients using quantization function $\gquant$: $\nabla_{W_l}^{q} = \gquant \left( \nabla_{W_l} \right)$. 
In this work, we aim to reduce the precision requirement for both weight gradients and activation gradients in the backward pass.


\setlength{\belowdisplayskip}{3pt}
\setlength{\belowdisplayshortskip}{3pt}
\setlength{\abovedisplayskip}{3pt}
\setlength{\abovedisplayshortskip}{3pt}

\section{Multiplicative Weight Update Algorithm for LNS}
\label{sec:quantized_weight_update}
\label{sec:lns-madam}

Although logarithmic quantized training significantly improves training efficiency, its overall efficiency continues to be hampered by the high precision requirement of weight updates.
We note that quantized weight update is orthogonal to quantized training due to the difference in their objectives. 
Quantized training tries to maintain the fidelity of weight gradients while accelerating forward and backward propagation. This provides accurate gradient information for the weight update. 
On the other hand, after receiving quantized weight gradients, quantized weight update aims to reduce gaps between updated weights and their (rounded) quantized counterparts.
Fig. \ref{fig:framework} distinguishes the two parts by different colors.

Previous works generally assume that the weight update is computed over a full-precision weight space. In other words, a full-precision copy of weights is maintained \cite{conv_log,otherlog} and very little rounding follows weight update. However, this offsets the efficiency benefits of quantized training and requires expensive floating-point arithmetic not available especially in cheap energy-constrained edge devices.
Therefore, in this work, we consider quantized weight update in LNS, where the weights are updated over a discrete logarithmic space instead of a full-precision one.
We aim to minimize the rounding error given that weights are represented in LNS.

\subsection{Quantized Weight Update}
To better understand this problem, we first define a generalized form of a weight update as: $W_{t+1} = U\left(W_{t}, \nabla_{W_{t}}\right),$ where $U$ represents any learning algorithm. For example, gradient descent (GD) algorithm takes $U_{GD} = W_t - \eta \, \nabla_{W_t},$ where $\eta$ is learning rate.

Because the weights need to be represented in a quantized format in LNS, it is necessary to consider the effect of logarithmic quantization during weight update. We define \textit{logarithmic quantized weight update} as follows:
\begin{equation}
\label{eq:quant_weight_update}
    W_{t+1}^{U} = \logquant \left(U\left(W_{t}, \nabla_{W_{t}}\right)\right).
\end{equation}
In this case, $W_{t+1}^{U}$ can be directly stored in a logarithmic format without using floating-point data type. For simplicity, we assume weight gradients $\nabla_{W_{t}}$ are exact as quantized training is orthogonal to this problem. Switching to the approximated gradient estimates will not affect our theoretical results.

\subsection{Quantization Error Analysis}
Because the logarithmic quantization requires representing values in a discrete logarithmic scale, quantized weight update inevitably introduces a mismatch between the quantized weights and their full-precision counterparts. To preserve the reliability of the optimization, we aim to reduce the quantization error (i.e., the mismatch). For the following, we take a theoretical perspective to discuss how different learning algorithms affect the quantization error under LNS. Detailed assumptions and proofs can be found in Appendix~\ref{apdx:analysis}.

Due to the logarithmic structure, we focus on minimizing a quantizaion error $r_{t} = \normsq{\log_2 \abs{W_{t+1}^{U}}  - \log_2 \abs{W_{t+1}}}$, which measures the L2 norm of the difference between the weights and their quantized counterparts in logarithmic space. Because $r_{t}$ quantify the relative difference between $\abs{W_{t+1}^{U}}$ and $\abs{W_{t+1}}$, minimizing $r_{t}$ is largely similar to minimizing the relative quantization error $\normsq{(W_{t+1} - W_{t+1}^{U}) / W_{t+1}}$.

We assume a simplified logarithmic quantization where the scale factor and the clamping function are ignored. This ensures our focus is on the effect of the quantization gap determined by $\gamma$ instead of the dynamic range. We also replace the deterministic rounding with a stochastic counterpart $SR$ where $\Expect \SR(x) = x$ for any real number. Although $SR$ helps us establish the theoretical results, in practice $SR$ requires random generators that induce additional costs, and thus are not suitable for energy-efficient training.
\begin{figure}[t]
    \centering
    \includegraphics[scale=0.35]{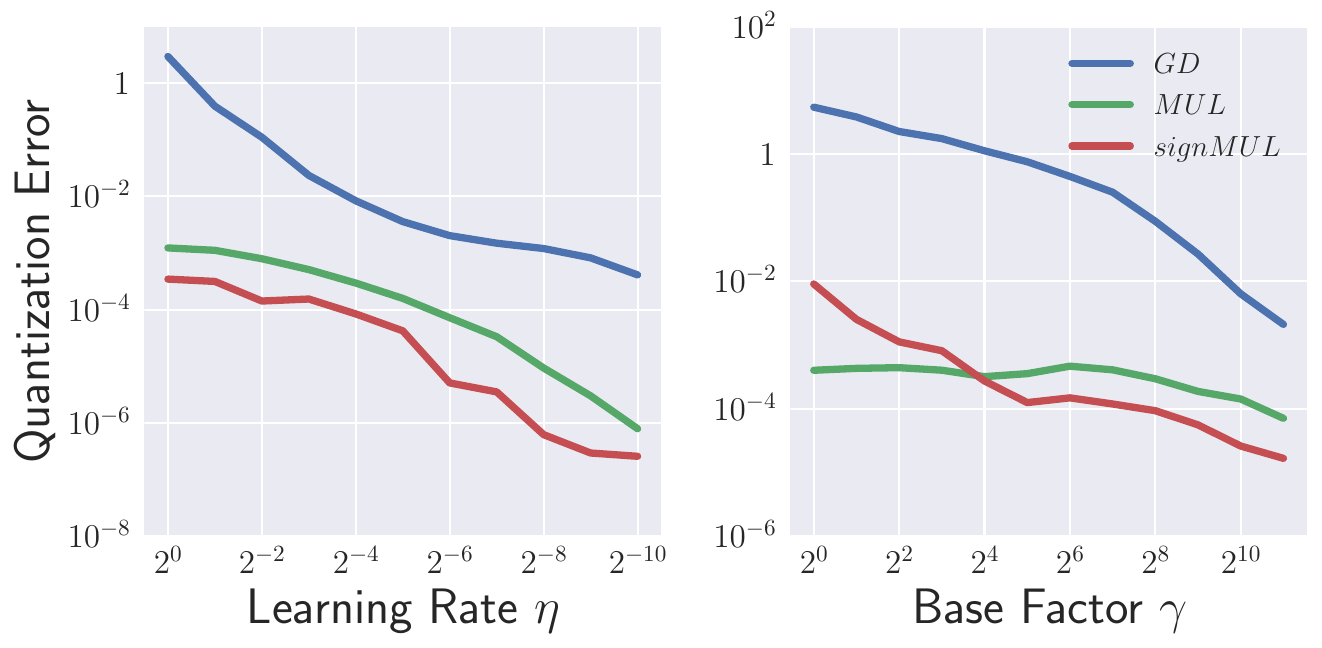}
    \vskip -0.1in
    \caption{Quantization error from different learning algorithms on ImageNet. The errors are averaged over all iterations in the first epoch. The results suggest that multiplicative algorithms introduce significantly lower errors compared to the gradient descent, which are also in line with our theoretical results.}
    \label{fig:quant_error}
\end{figure}

Given everything we need, we use gradient descent as an example to discuss why traditional learning algorithms are not suited for LNS-based quantized weight updates. The theorem is stated as follows:
\begin{restatable}{theorem}{errorGD}\label{thm:errorGD}
    The quantization error $r_{t,GD}$ introduced by logarithmic quantized gradient descent at iteration $t$ can be bounded in expectation, as:
    \begin{equation}
        \Expect r_{t,GD} \leq \frac{\sqrt{d}}{\gamma} \, \norm{\log_2 \left( \abs{W_{t}} - \eta_1 \nabla_{W_{t}} \right)},
        \label{eq:error_gd}
    \end{equation}
    where d is the dimension of $W$ and $\eta_1$ is the learning rate of $U_{GD}$.
\end{restatable}

Theorem \ref{thm:errorGD} suggests that $r_{t,GD}$ is magnified when the magnitudes of weights become larger. This is because the updates $\eta_1 \nabla_{W_{t}}$ generated by GD are not proportional to the magnitudes of weights. $\eta_1 \nabla_{W_{t}}$ can be orders of magnitude smaller than the quantization gaps as weights become larger, and thus these updates often are disregarded by quantization function $\logquant$. We intuitively illustrate this problem in Fig. \ref{fig:add_vs_mul}.

To ensure the updates are proportional to the weights, a straightforward way is to update the weights multiplicatively. Because the weights are represented in LNS, we further consider a special multiplicative learning algorithm tailored to LNS, which updates the weights directly over their logarithmic space:
\begin{equation}
    U_{MUL} = sign(W_t) \odot 2^{\Tilde{W_t} - \eta \, \nabla_{W_t} \odot sign(W_t)}
\end{equation}
where $\Tilde{W_t}=\log_2 \abs{W_t}$ are the exponents of the magnitude of weights, and $\odot$ denotes element-wise multiplication. $U_{MUL}$ makes sure the magnitude of each element $W_t(k)$ of the weights decreases when the sign $sign(W_t(k))$ and $\nabla_{W_t(k)}$ agree and increases otherwise. The quantization error with regards to $U_{MUL}$ is stated as follows:

\begin{restatable}{theorem}{errorMUL}\label{thm:errorMUL}
    The quantization error $r_{t,MUL}$ introduced by logarithmic quantized multiplicative weight update at iteration $t$ can be bounded in expectation, as:
    \begin{equation}
        \Expect r_{t,MUL} \leq \frac{\sqrt{d}\, \eta_2}{\gamma} \, \norm{ \nabla_{W_{t}}},
    \end{equation}
    where d is the dimension of $W$ and $\eta_2$ is the learning rate of $U_{MUL}$.
\end{restatable}

Theorem \ref{thm:errorMUL} indicates that $r_{t,MUL}$ does not depend on the magnitudes of weights, and thus the quantization error is not magnified when the weights become larger. 
This is in stark contrast to the quantization error from gradient descent shown in Equation~\ref{eq:error_gd}. 
The comparison is illustrated in Fig. \ref{fig:add_vs_mul}.

Interestingly, we find that the quantization error $r_{t,MUL}$ can be further simplified by regularizing the information of gradients for the learning algorithm $U_{MUL}$:
\begin{restatable}{lemma}{errorMULsign}\label{lem:errorMULsign}
    Assume the multiplicative learning algorithm $U_{MUL}$ only receives the sign information of gradients where $U_{MUL} = \Tilde{W}_t - \eta_2 \, sign(\nabla_{W_t}) \odot sign(W_t)$. The upper bound on quantization error $r_{t,MUL}$ becomes:
    \begin{equation}
        \Expect r_{t,MUL} \leq \frac{d \, \eta_2}{\gamma}.
    \end{equation}
\end{restatable}
The result in Lemma \ref{lem:errorMULsign} suggests that $r_{t,MUL}$ can be independent of both weights and gradients when only taking sign information of gradients during weight update. We denote this special learning algorithm as $U_{signMUL}$. $U_{signMUL}$ is a multiplicative version of signSGD, which has been studied widely \cite{bernstein2018signsgd}. 

To verify our theoretical results, we empirically measure the quantization errors for the three aforementioned learning algorithms over a range of $\eta$ and $\gamma$. As shown in Fig. \ref{fig:quant_error}, the empirical findings are in line with our theoretical results. Although all learning algorithms introduce less errors when $\eta$ and $\gamma$ become smaller, the multiplicative algorithms introduce significantly lower errors compared to the gradient descent. 

In addition to reducing the quantization error in the quantized weight update, $U_{signMUL}$ must also have the ability to minimize the loss function $\loss(W)$. Interestingly, we notice that $U_{signMUL}$ resembles a recently proposed learning algorithm Madam, where Bernstein et al. \cite{bernstein2020learning} proves that Madam optimizes the weights in a descent direction.
Madam updates the weights multiplicatively using normalized gradients:
\begin{equation}\label{eq:madam}
\begin{split}
    U_{\textit{Madam}} &= W_t \odot e^{-\eta \,sign(W_t) \odot\, g_{t}^{*}} \\
    g_{t}^{*} &= g_{t} / \sqrt{g_{2_{t}}}
\end{split}
\end{equation}
where $g_{t}$ represents the gradient vector $\nabla_{W_t}$, and $g_{t}^{*}$ denote a normalized gradient, which is the fraction between $g_{t}$ and the square root of its second moment estimate $\sqrt{g_{2_{t}}}$. Bernstein et al. \cite{bernstein2020learning} demonstrates that Madam achieves state-of-the-art accuracy over multiple tasks with a relatively fixed learning rate $\eta$. They also theoretically prove the descent property of Madam that ensures its convergence. Although Bernstein et al. \cite{bernstein2020learning} further shows the possibility of applying Madam over a discrete logarithmic weight space, they still employ full-precision training without considering low-precision LNS. 
\begin{figure*}[tbh]
    \centering
    \centerline{\includegraphics[scale=0.5]{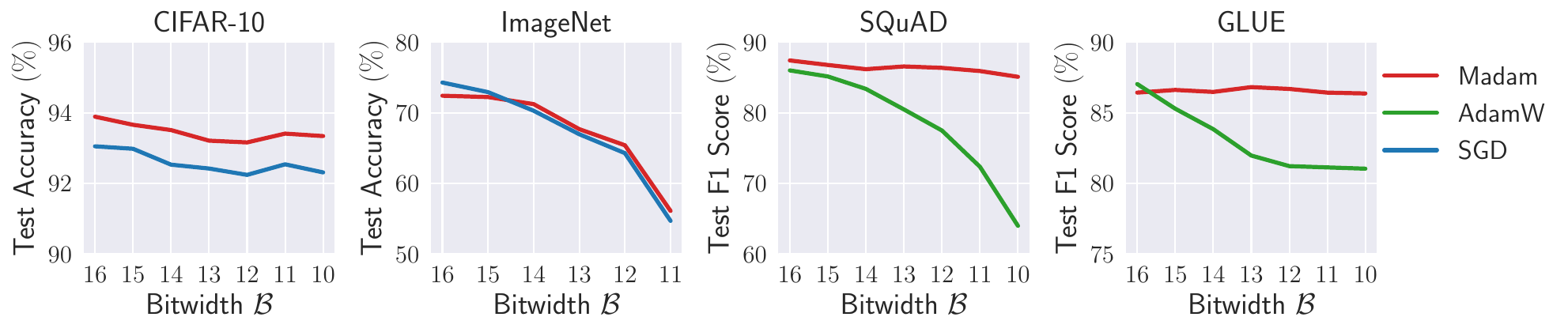}}
    \vskip -0.15in
    \caption{Comparing Madam with SGD and Adam optimizers under the logarithmic quantized weight update (defined in Equation \ref{eq:quant_weight_update}). The bitwidth of the weight update $Q_{U}$ is varied from 16-bit to 10-bit.}
    \label{fig:LNS-Madam}
    \vskip -0.2in
\end{figure*}


To ensure low-precision weight updates in LNS, we apply a modified version of the Madam optimizer to enable fast convergence while preserving low quantization error. The modified Madam directly optimizes the weights over their base-2 logarithmic space using the gradient normalization technique described in Equation \ref{eq:madam}. Details of our optimizer are shown in Algorithm \ref{alg:lns_madam}.
Because our Madam optimizer directly updates base-2 exponents of weights in LNS, there is no need for integer-to-LNS conversion during weight update when the weights are already in LNS, further reducing the energy cost.

\section{Hardware Implementation}
\label{sec:hardware}

\begin{figure}[t]
    \centering
    \includegraphics[trim=0 83 600 0, clip,width=0.93\columnwidth]{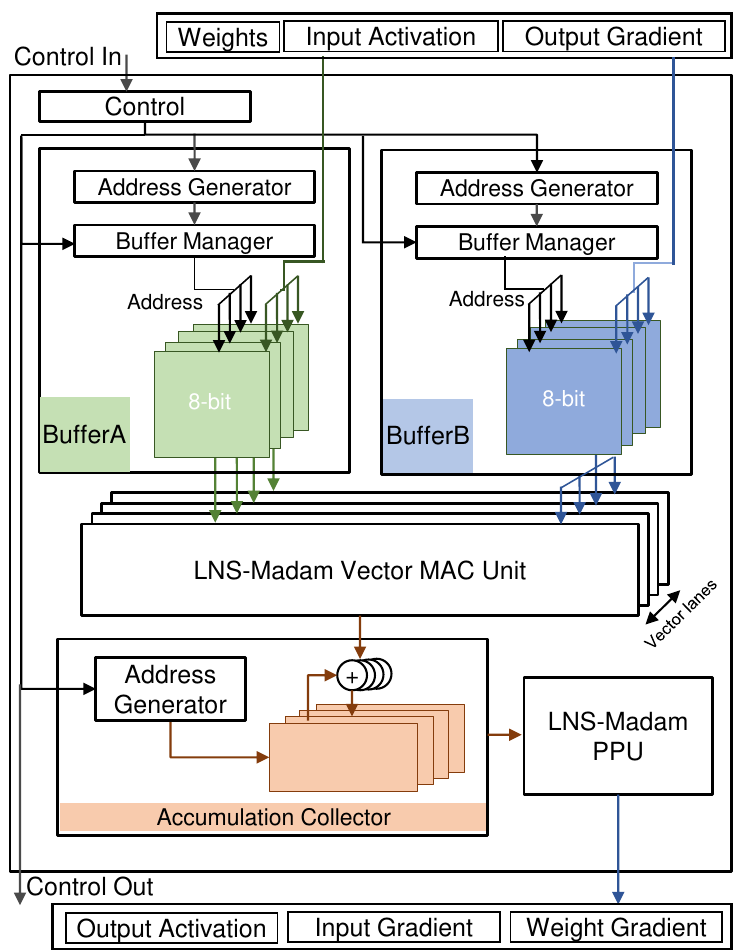}
    \caption{LNS-Madam processing element (PE).}
    \label{fig:lns-pe}
\end{figure}

We extend a previously optimized DNN accelerator \cite{magnet} to support LNS-based DNN computations.
Fig.~\ref{fig:lns-pe} shows the micro-architecture of the PE which performs dot-product operations. 
Each PE consists of set of vector MAC units fed by the buffers that store weights, input activations, and output gradients.
Additionally, the accumulation collectors store and accumulate partial sums which are passed to the PPU for post-processing (e.g., quantization scaling, non-linear activation functions) if necessary.

\begin{figure*}[t]
    \centering
    \centerline{\includegraphics[trim=0 223 30 0, clip,width=1.9\columnwidth]{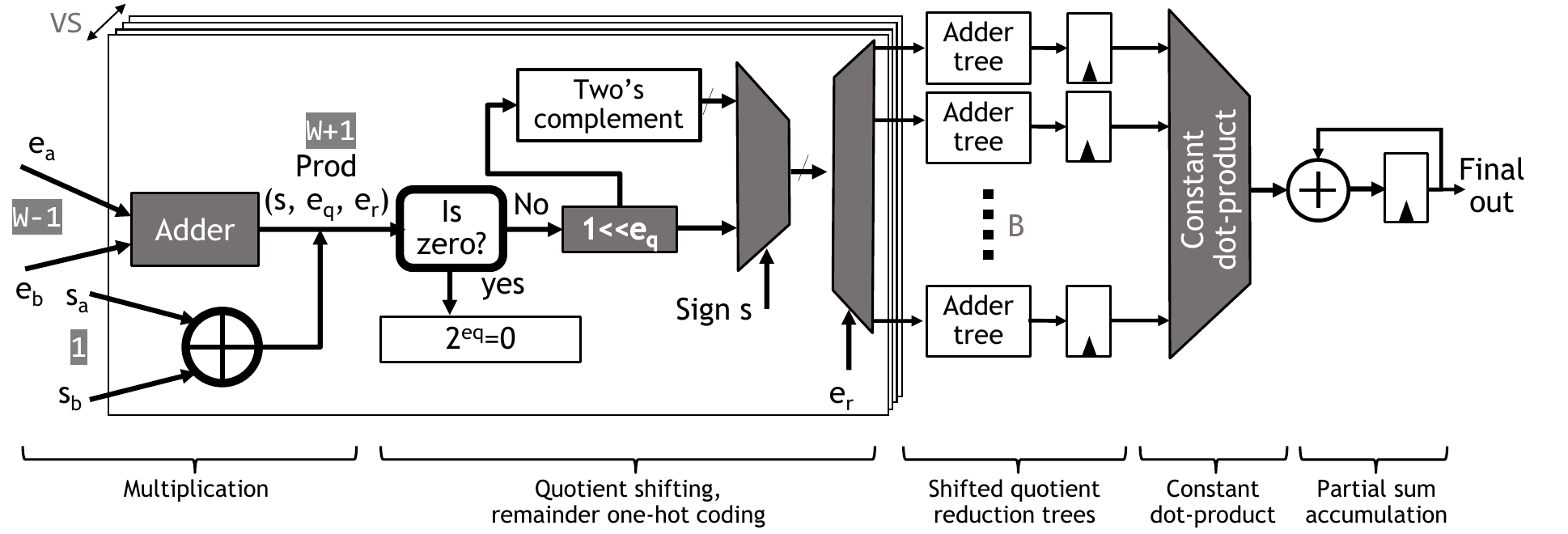}}
    \vskip -0.1in
    \caption{LNS-Madam Vector MAC Unit -- Performs dot-products of inputs represented in LNS and produces partial sum outputs in integer format. Bitwidths of different signals are highlighted. \TT{VS} stands for vector size; \TT{W} stands for bitwidth of input values; \TT{B} refers to base factor and number of remainder bins.}
    \label{fig:lns-datapath}
    \vskip -0.1in
\end{figure*}

Fig.~\ref{fig:lns-datapath} shows the LNS-based datapath inside the LNS-Madam Vector MAC Unit.
Here we model exact LNS-to-integer conversion without any approximation.
With a vector size of 32 and input bitwidths of 8, the datapath processes 32 7-bit exponent values at each of its exponent inputs ($e_a$ and $e_b$) and 32 1-bit sign values ($s_a$ and $s_b$) at each of its sign inputs to produce a 24-bit partial sum.
First, the LNS datapath performs the dot-product multiplication by adding the exponents and XOR-ing the sign bits.
The output of the product computation requires an additional bit to account for the carry-out of the adder.
At this point, each exponent is split into a quotient component ($e_q$) and a remainder component ($e_r$) based on the LSB/MSB property mentioned in Section~\ref{subsec:log_operations}.
Second, the datapath performs shifting by the quotient to implement the quotient component in Equation~\ref{eq:decomposition}.
Depending on the sign bit, the corresponding signed shifted value is selected and passed to the corresponding adder tree based on the remainder select signal.
Third, the result of shifted values are reduced through the set of adder trees and registered.
At last, the results of the adder trees are multiplied with corresponding remainder constants (described in Section~\ref{subsec:log_operations}) from a LUT and accumulated into the final partial sum, represented in integer (linear) format.
This partial sum needs to be converted back into logarithmic format and written back to the global buffer for subsequent LNS-based computations.

Additional microarchitectural details of the PE are listed in Table~\ref{tab:microarchitecture_details}.
Notably, our accelerator uses a multi-level dataflow called output-stationary local-A-stationary\cite{magnet} to optimize reuse across different operands. Inputs from buffer A are read out once every 16 cycles and stored in a register for temporal reuse. Inputs from buffer B are read once every cycle and reused across the 32 lanes spatially. Partial sums are temporally accumulated in a 16-entry latch array collector before sending the completed sum to the post-processing unit.
The two buffers in the PE store different data depending on whether output activation, input gradient, or weight gradient is being computed.
For example, weights and input activations are stored in \TT{BufferA} and \TT{BufferB} respectively during forward propagation to compute the output activations.
On the other hand, input activations and output gradients are stored in the respective buffers during backward propagation to compute the weight gradient.
Table~\ref{tab:dataflow} outlines how we map various tensors in DNN computation to buffers in our hardware during different computation passes.
Note that weight updates are performed outside of the PEs through the global buffer.

\begin{table}[tbh]
\label{tab:microarchitecture_details}
\begin{center}
\caption{Microarchitectural details of LNS-Madam PE}
\begin{tabular}{c|c}
\hline \hline
Dataflow & Multi-level \\
\hline
Vector size / \# Vector lanes &  32 \\
\hline
Weight/activation precision & 8-bit \\
\hline
Gradient precision & 8-bit \\
\hline
\# Remainder Bins & 8 \\
\hline
Accumulation precision &  24-bit \\
\hline
Accumulation collector size & 1.5 KB \\
\hline
BufferA size & 128 KB \\
\hline
BufferB size & 8 KB\\
\hline
\end{tabular}
\end{center}
\end{table}

\begin{table}[tbh]
    \begin{center}
    \begin{threeparttable}
        \caption{Mapping of tensors to buffers in PE during different computation passes\tnote{1}}
        \label{tab:dataflow}
        \begin{tabular}{ccc}
        \toprule
        \textbf{Pass} & 
        \textbf{BufferA} &
        \textbf{BufferB} \\ 
        \midrule
        Forward & Weight & Input Activation \\ 
        Backward (Input) & Weight & Output Gradient \\
        Backward (Weight) & Input Activation & Output Gradient \\
        \bottomrule
        \end{tabular}
        \begin{tablenotes}
        \item[1] Backward pass consists of backward computation for input gradient, denoted \TT{Backward(Input)}, and backward computation for weight gradient, denoted \TT{Backward(Weight)}.
        \end{tablenotes}
        \end{threeparttable}
        \end{center}
    \end{table}

\section{Experiments}
\label{sec:experiments}

In this section, we evaluate both the accuracy and energy efficiency of using LNS-Madam to train state-of-the-art models on large-scale datasets.

\subsection{Model Accuracy}

To evaluate accuracy, we simulate LNS-Madam using a PyTorch-based neural network quantization library that implements a set of common neural network layers (e.g., convolution, fully-connected) for training and inference in both full and quantized modes \cite{wu2020integer}.  
The baseline library supports integer quantization in a fixed-point number system, and we further extend it to support LNS. The library also provides utilities for scaling values to the representable integer range of the specific number format.
With this library, a typical quantized layer consists of a conventional layer implemented in floating-point preceded by a weight quantizer and an input quantizer that converts the weights and inputs of the layer to the desired quantized format. For the backward pass, after the gradients pass through the STE in each quantizer, they will be quantized by their quantizers as well.

We benchmark LNS-Madam on various tasks including ResNet models on CIFAR-10 and ImageNet, and BERT-base and BERT-large language models on SQuAD and GLUE. Specifically, we train ResNet models from scratch on CIFAR-10 and ImageNet, and fine-tune pre-trained BERT models on SQuAD and GLUE. Detailed descriptions of datasets and models can be found in the appendix.

\begin{table}[t!]
\centering
\begin{threeparttable}
\caption{Base Factor Selection on ImageNet\tnote{1,2}}
\label{tab:base_factor_selection}
\setlength{\tabcolsep}{11pt}
    \begin{tabular}{cccc}
    \toprule
    \textbf{$\gamma$} & \textbf{Dynamic Range}& \textbf{Forward} & \textbf{Backward}
    \\
    \midrule
        \textbf{$1$
    } &   (0,127)  & \text{NaN}  & \text{NaN}  \\
        \textbf{$2$
    } &   (0,63.5)  & 75.81   & 75.79  \\
        \textbf{$4$
    } &   (0,31.8)  & 75.96   & 76.07  \\
        \textbf{$8$
    } &   (0,15.9)  & 75.88   & 76.23  \\
        \textbf{$16$
    } &   (0,7.9)  & 76.32  & 63.67  \\
        \textbf{$32$
    } &   (0,4.0)  & 68.15  & 20.71  \\
    \bottomrule
    \end{tabular}
    \begin{tablenotes}
    \item[1] Bitwidth is 8-bit across settings. Quant Forward or Quant Backward denotes the settings where either forward propagation or backward propagation is quantized while leaving the rest of computation in full-precision. 
    \item[2] The results of test accuracy (\%) are listed.
    \end{tablenotes}
\end{threeparttable}
\end{table}


\subsubsection{Parameter Settings} 
We fix the bitwidth to be 8-bit for both forward and backward propagation, which includes the bitwidth for weights, activations, activation gradients, and weight gradients. We note that 8-bit weight gradients are lower than previous studies that use 16-bit or even 32-bit weight gradients \cite{4_bit_ultra_low,conv_log}.

To find an appropriate base factor $\gamma$ under the 8-bit setting, we vary $\gamma$ to find the appropriate dynamic ranges for forward and backward qantization. The dynamic range in LNS is $(0, (2^{\bitwidth -1} - 1) / \gamma)$, which is controlled by both bitwidth and base factor. 

As shown in Table \ref{tab:base_factor_selection}, we fix the bitwidth as 8-bit and vary the base factor $\gamma$ to find the appropriate dynamic ranges for forward and backward quantization. According to the results, we find the base factor of 8 with the dynamic range $(0,15.9)$ that uniformly works across $\wquant$,$\aquant$,$\equant$, and $\gquant$. 

In order to maintain optimization stability, the bitwidth of the weight updates require to be larger than the bitwidth of the weights. When the bitwidth of $\uquant$ is larger than 8-bit, we increase its base factor to match the desired dynamic range $(0,15.9)$.

We empirically search the best learning rate $\eta$ for our Madam optimizer from $2^{-4}$ to $2^{-10}$, and we find $\eta=2^{-7}$ works best uniformly across tasks, which suggests the learning rate for Madam is robust.

\begin{table}[t!]

    \centering
    
    \begin{threeparttable}
    \caption{Benchmarking LNS-Madam on various datasets and models\tnote{1}}
    \label{tab:standard_benchmark}
    \setlength{\tabcolsep}{6pt}
        \begin{tabular}{ccccc}
        \toprule
        \textbf{Dataset} & 
        \textbf{Model} &
        \textbf{LNS-Madam\tnote{2}} &
        \textbf{FP8\tnote{2}} &
        \textbf{FP32}
        \\
        \midrule
        CIFAR-10  & ResNet-18   & $93.41$ & $93.12$ & $93.51$ \\
        ImageNet  & ResNet-50   & $76.14$ & $75.83$ & $76.38$ \\
        SQuAD     & BERT-base   & $88.13$ & $88.07$ & $88.36$ \\
        SQuAD     & BERT-large  & $90.75$ & $90.54$ & $90.80$ \\
        GLUE      & BERT-base   & $88.89$ & $88.73$ & $88.92$ \\
        GLUE      & BERT-large  & $89.24$ & $88.91$ & $89.35$ \\
        \bottomrule
        \end{tabular}
        \begin{tablenotes}
            \item[1] The results of test accuracy (\%) are listed. \\
            \item[2] Forward and backward propagation are in 8-bit, and the weight update is in 16-bit.
        \end{tablenotes}
    \end{threeparttable}

    \end{table}
\begin{table}[t!]

    \centering
    \begin{threeparttable}
    \caption{Comparing LNS-Madam with recent low-precision training methods on 8-bit training\tnote{1}}
    \label{tab:compare_recent_methods}
    \setlength{\tabcolsep}{7.5pt}
        \begin{tabular}{cccc}
        \toprule
        & \textbf{Data format} & \textbf{16-bit} & \textbf{32-bit}  \\
        \midrule
        LNS-Madam & LNS & \textbf{76.14} & 76.23 \\
        BHQ \cite{chen2020statistical} & INT &  74.89 & \textbf{76.35} \\
        Unified INT8 \cite{zhu_towards_2020} & INT & 74.73 & 76.27 \\
        FP8 \cite{8bit} & FP & 71.46 & 71.53 \\
        \bottomrule
        \end{tabular}
        \begin{tablenotes}
            \item[1] Evaluate ResNet-50 on ImageNet. Forward and backward propagation are in 8-bit. Test accuracy (\%) evaluated under 32-bit and 16-bit weight update are presented. \\
        \end{tablenotes}
    \end{threeparttable}
\end{table}


\begin{table}[t!]

    \centering
    \begin{threeparttable}
    \caption{Comparing LNS-Madam and BHQ over a range of bitwidth\tnote{1}}
    \label{tab:compare_bhq}
    \setlength{\tabcolsep}{6pt}
        \begin{tabular}{cccccc}
        \toprule
        & \textbf{4-bit} & \textbf{5-bit} & \textbf{6-bit} & \textbf{7-bit} & \textbf{8-bit}  \\
        \midrule
        LNS-Madam & \textbf{74.23} & \textbf{75.89} & 74.41 & \textbf{76.16} & 76.23 \\
        BHQ \cite{chen2020statistical} & 74.04 & 75.70 & \textbf{76.21} & 76.14 & \textbf{76.35} \\
        \bottomrule
        \end{tabular}
        \begin{tablenotes}
            \item[1] Bitwidth of activation gradients varies from 4-bit to 8-bit. The results of test accuracy (\%) are listed. \\
        \end{tablenotes}
    \end{threeparttable}
\end{table}
\subsubsection{Comparisons}
Given the settings above, we compare LNS-Madam with FP8 and FP32. For FP8 and FP32, the standard optimizers are applied for tasks by default. We use a tuned SGD optimizer for CIFAR-10 and ImageNet datasets, and a tuned AdamW optimizer for SQuAD and GLUE datasets. For all settings, we use per-channel scaling for ResNet and per-feature scaling for BERT, both of which are commonly used scaling techniques. The clamping function is performed by matching the largest value within each group of numbers.

To demonstrate LNS-Madam is better than popular number systems, we compare it with both FP8 and FP32, where our FP8 contains 4-bit exponent and 3-bit mantissa. As shown in Table \ref{tab:standard_benchmark}, LNS-Madam yields better performance than FP8, and it even achieves performance comparable to the full-precision counterpart.

In addition, we compare LNS-Madam with recent methods on low-precision training. For all methods, we fix forward and backward propagation in 8-bit while varying the weight update precision from 32-bit to 16-bit. As shown in Table~\ref{tab:compare_recent_methods}, LNS-Madam achieves the best accuracy under 16-bit weight update, which demonstrates its effectiveness under the quantized setting. FP8 \cite{8bit} also achieves negligible degradation after switching to 16-bit, as it applies stochastic rounding over weight update process. Since BHQ \cite{chen2020statistical} achieves the best accuracy under 32-bit, we also compare it with LNS-Madam over a range of bitwidth settings in Table \ref{tab:compare_bhq}.

We also compare Madam with the default optimizers SGD and AdamW under logarithmic quantized weight update, as defined in Equation \ref{eq:quant_weight_update}. All optimizers use the same learning rates as above. 

As shown in Fig. \ref{fig:LNS-Madam}, we vary the bitwidth of the quantized weight update $Q_U$ from 16-bit to 10-bit to test their performance over a wide range. The results suggest compared to other optimizers, \emph{Madam always maintains higher accuracy when precision is severely limited}. Notably, for BERT model on SQuAD and GLUE benchmarks, Madam is 20\% better than Adam with respect to F-1 score, when the weight update is in 10-bit. We observe large degradation for both Madam and SGD on ImageNet training, and we believe this is because the weights in some layers inevitably require higher precision settings. We leave it as future work to explore LNS-Madam under a customized precision setting.

\begin{table}[H]
\label{tab:experimental_setup}
\begin{center}
\caption{Design tools used for LNS-Madam hardware experiments}
\begin{tabular}{c|c}
\hline \hline
HLS Compiler & Mentor Graphics Catapult HLS \\
\hline
Verilog simulator & Synopsys VCS \\
\hline
Logic synthesis & Synopsys Design Compiler \\
\hline
Place-and-route & Synopsys ICC2 \\ \hline
Power Analysis & Synopsys PT-PX \\
\hline \hline
\end{tabular}
\end{center}
\end{table}

\subsection{Energy Efficiency}
\label{sec:energy_analysis}
We leverage the hardware implementation described in Section~\ref{sec:hardware} to evaluate the energy efficiency of LNS-Madam. 
We code  the hardware model in C++ and Verilog and synthesize it to a combined cycle-accurate RTL using a commercial high-level synthesis tool~\cite{mcfarland1990high}.
Once the RTL is generated, a standard logic synthesis flow is used to obtain the gate-level netlist that is then simulated with representative inputs. 
To extract energy consumption, we supply the gate-level simulation results from post-synthesis to a standard power analysis tool. We then use an analytical model to compute the total energy consumption for different workloads.
We perform our analysis in a sub-16nm state-of-the-art process technology at 0.6V targeting a frequency of 1.05 GHz. 
Table~\ref{tab:experimental_setup} summarizes the design tools used in the evaluation. 

\begin{table}[H]
    \centering
     \vskip -0.05in
  \setlength{\tabcolsep}{10pt}
  \begin{threeparttable}
  \caption{Energy efficiency for different models and number formats\tnote{1}}
  \label{tab:energy_analysis}
  \begin{tabular}{ccccc}
      \toprule
      \textbf{Model} &
      \textbf{LNS}  &
      \textbf{FP8} &
      \textbf{FP16} &
      \textbf{FP32} \\
      \midrule
      ResNet-18 & 0.54 & 1.22 & 2.50 & 5.99 \\
      ResNet-50 & 0.99 & 2.25 & 4.59 & 11.03 \\
      BERT-Base & 7.99 & 18.23 & 37.21 & 89.35 \\
      BERT-Large & 27.85 & 63.58 & 129.74 & 311.58 \\
      \bottomrule
      
    \end{tabular}
      \begin{tablenotes}
      \item[1] The per-iteration energy consumption in mJ are listed.
      \end{tablenotes}
    \end{threeparttable}
     \vskip -0.25in
  \end{table}
\begin{figure}[h]
    \centering
    \includegraphics[scale=0.55]{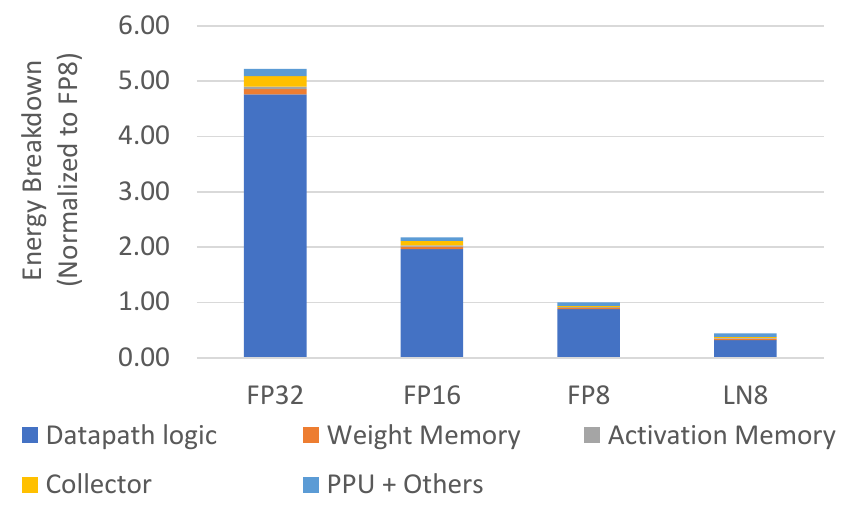}
    \caption{Energy breakdown of the PE shown in Fig. \ref{fig:lns-datapath} for different dataformats.}
    \label{fig:energy_breakdown}
    \vskip -0.25in
\end{figure}

\begin{figure}[h]
    \centering
    \includegraphics[scale=0.7]{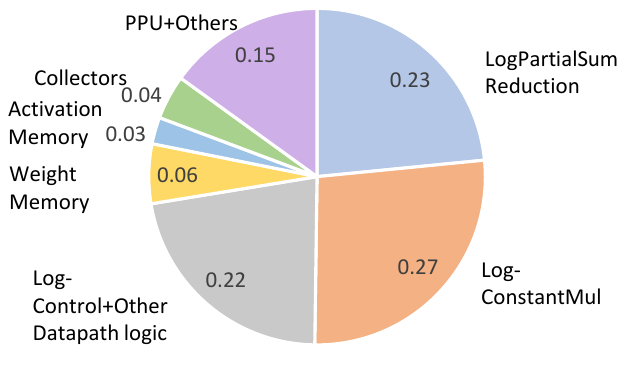}
    \caption{LNS PE energy breakdown showing different components of datapath}
    \label{fig:energy_breakdown_ln8}
\end{figure}
\begin{table*}[t]

    \centering
    \begin{threeparttable}
    \caption{Comparing LNS-Madam with recent LNS-based designs}
    \label{tab:lns_related_works}
    \setlength{\tabcolsep}{5pt}
        \begin{tabular}{cccccc}
        \toprule
        & \textbf{LNS-Madam} & Arnab et al. \cite{sanyal_neural_2020} & Miyashita et al. \cite{conv_log} & Lee et al. \cite{lognet} & Vogel et al. \cite{otherlog} \\
        \midrule
        Support inference or fine-tuning?          & \cmark & \cmark & \cmark & \cmark & \cmark \\
        Support training from scratch?             & \cmark & \cmark & \xmark & \xmark & \xmark \\
        Weight update precision                    & \textless 16-bit & 32-bit  & 32-bit & 32-bit  & 32-bit \\
        Efficient log-to-linear conversion support & \cmark & \cmark & \cmark & \cmark & \xmark \\
        Large-scale evaluation                     & \cmark & \xmark & \cmark & \cmark & \cmark \\
        \bottomrule
        \end{tabular}
    \end{threeparttable}
\end{table*}
\begin{figure}[H]
    \centering
    \includegraphics[scale=0.4]{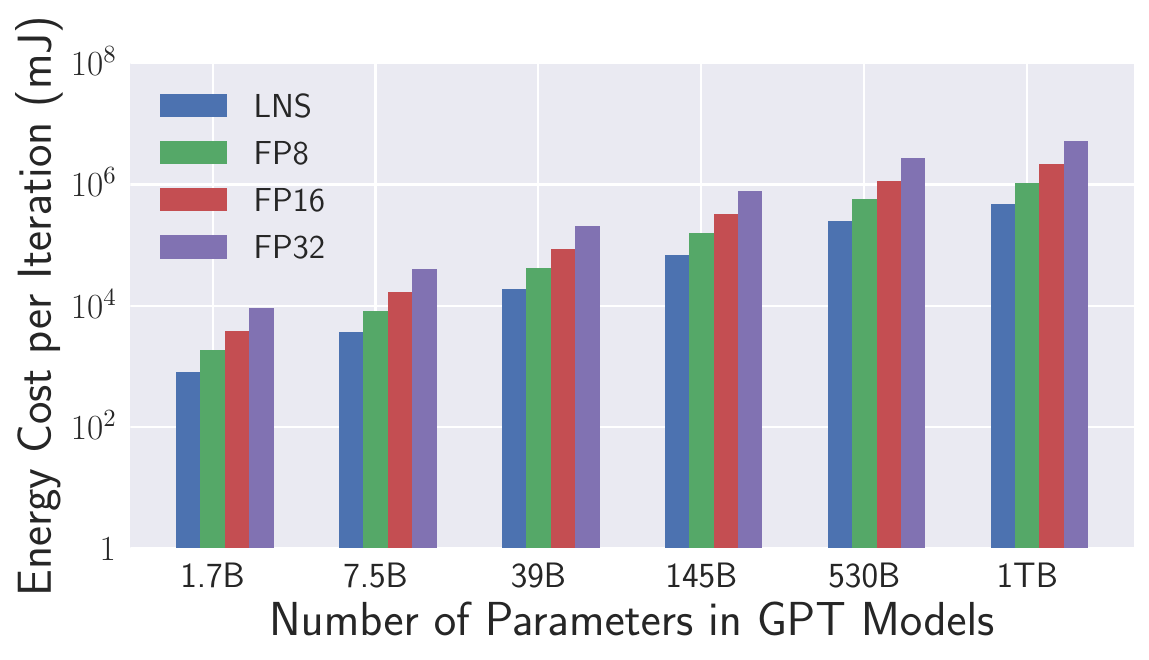}
    \caption{Energy efficiency over a range of GPT models from 1 billion to 1 trillion parameters. The models are scaled by a throughput efficient method proposed by Narayanan et al. \cite{narayanan2021efficient}.}
    \label{fig:energy_gpt}
\end{figure}

In our experiment, the LNS-Madam hardware is designed with bitwidth $\bitwidth = 8$ and base factor $\basefactor=8$ for both forward and backward computations. 
In addition to experimenting with the LNS-based datapath shown in Fig.~\ref{fig:lns-datapath}, we also consider FP8, FP16, and FP32 datapath baselines for comparison. 

Table \ref{tab:energy_analysis} presents the energy efficiency per iteration of one forward pass and one backward pass of training.
Because different number systems share the same training iterations, per-iteration energy results imply the energy comparison over the entire training.
We also present the energy breakdown of the whole PE in Figure \ref{fig:energy_breakdown}. As shown in the figure, FP arithmetic is extremely expensive, contributing a large fraction to the total energy consumption of PEs. The proposed LNS datapath offers significant reduction in the logic complexity, leading to 2.2X, 4.6X, and 11X energy efficiency improvements over FP8, FP16, and FP32 implementations, respectively. We also provide a detailed energy breakdown showing different components of the LNS PE in Fig. \ref{fig:energy_breakdown_ln8}.
In addition, Fig. \ref{fig:energy_gpt} shows the energy efficiency over a range of GPT models from 1 billion to 1 trillion parameters. 

\section{Related Works}

\subsection{Low-precision training}
To achieve good accuracy at reduced precision, quantization-aware training (QAT) is commonly applied to directly train the quantized model using straight-through estimators \cite{bengio2013estimating,zhou2016dorefa,rastegari2016xnor,zhou2017incremental,jacob2018quantization}. To accelerate the training phase, several studies suggest quantizing the gradients during backward propagation \cite{scalable_8bit, 4_bit_ultra_low, 8bit}. To maintain the fidelity of the gradient accumulation, some low-precision training methods assume a full-precision copy for weights during the weight update \cite{scalable_8bit, chen2020statistical}. Other studies reduce the precision for the weight update by using high-precision gradient accumulator \cite{esakr2018pertensor}, stochastic rounding \cite{wage,8bit} or additionally quantizing the residual part of weights \cite{sun2019hybrid,desa2018highaccuracy}. Cambricon-Q accelerates the weight update from a hardware perspective by avoiding costly data transferring in weight update \cite{zhao_cambricon-q_2021}.
However, they mostly apply SGD or Adam during the weight update without considering the relationship between the precision of the weights and the underlying learning algorithms. 

\subsection{Logarithmic number system}
Previous works demonstrate the effectiveness of using logarithmic representation for DNNs \cite{conv_log,lognet, johnson2018rethinking, fb_log}. Furthermore, some studies suggest using multiple levels of log-base to reduce the quantization error \cite{conv_log,otherlog}. However, few of them address the additional computational cost induced by this multi-base design nor scale the system to state-of-the-art neural networks for both training and inference. From the perspective of hardware design, a few studies focus on improving the efficiency of LNS by utilizing the significant cost reduction of multiplications \cite{saadat2018minimally,saadat2019approximate,saadat2020realm,johnson2018rethinking, fb_log}. We compare LNS-Madam with recent LNS-based designs on Table~\ref{tab:lns_related_works}.

\subsection{Multiplicative weight update}
 Multiplicative algorithms, such as exponentiated gradient algorithm and Hedge algorithm in AdaBoost framework \cite{manfred,adaboost}, have been well studied in the field of machine learning. In general, multiplicative updates are applied to problems where the optimization domain's geometry is described by relative entropy, such as probability distribution \cite{manfred}. Recently, \cite{bernstein2020learning} proposes an optimizer Madam that focuses on optimization domains described by any relative distance measure instead of only relative entropy. Madam shows great performance in training large-scale neural networks. However, Madam requires full-precision training without considering its connection to LNS-based low-precision training.
\section{Conclusions}

In this work, we propose a co-designed low-precision training framework LNS-Madam that jointly considers the logarithmic number system and the multiplicative weight update algorithm. Experimental results show that LNS-Madam achieves comparable accuracy to full-precision counterparts even when forward and backward propagation, and weight updates are all in low-precision. 
To support the training framework in practice, we design a hardware implementation of LNS-Madam to efficiently perform the necessary LNS computations for DNN training. 
Based on our energy analysis, LNS-Madam reduces energy consumption by over 90\% compared to a floating-point baseline. 

An important application of our low-precision training framework is learning neural networks over energy-constrained edge devices. This is fundamental for intelligent edge devices to easily adapt to changing and non-stationary environments by learning on-device and on-the-fly. By enabling highly energy-efficient training, our work carries the promising opportunity for using LNS-based hardware to conduct environmental-friendly deep learning research in the near future.


\bibliographystyle{IEEEtran}
\bibliography{refs}



\newpage
\begin{IEEEbiography}
[{\includegraphics[width=1in,height=1.25in,clip,keepaspectratio]{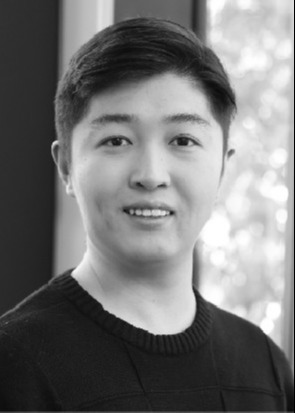}}]{Jiawei Zhao}
received the B.S. degree in Computer Science from NUAA, Nanjing, China, in 2019. He is currently pursuing the Ph.D. degree at Caltech, under the supervision of Prof. Anima Anandkumar. His research interest lies in machine learning, specifically in deep learning and optimization. He is working on the development of novel and efficient algorithms for solving deep learning problems, such as distributed or low-precision training.
\end{IEEEbiography}

\begin{IEEEbiography}
[{\includegraphics[width=1in,height=1.25in,clip,keepaspectratio]{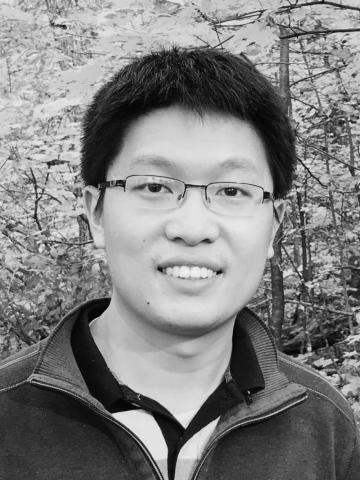}}]{Steve Dai}
is currently a Research Scientist as part of the ASIC \& VLSI Research Group at NVIDIA. His research interests include energy-efficient DL acceleration, high-level design methodologies, and ML-assisted EDA. Dai received the B.S. degree in electrical engineering from the University of California at Los Angeles in 2011, the M.S. degree in electrical engineering from Stanford University in 2013, and the Ph.D. degree in electrical and computer engineering from Cornell University in 2019.
\end{IEEEbiography}

\begin{IEEEbiography}
[{\includegraphics[width=1in,height=1.25in,clip,keepaspectratio]{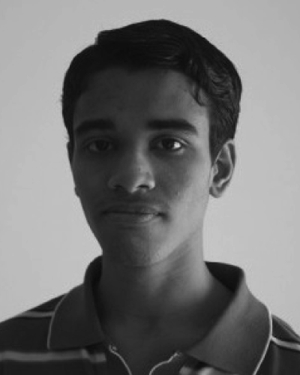}}]{Rangharajan Venkatesan}
is a Senior Research Scientist with NVIDIA. His research interests include machine learning  accelerators, low-power VLSI design, and SoC design methodologies. He has served as a member of the technical program committees of several leading IEEE conferences including International Solid-State Circuits Conference,International  Symposium  on  Micro architecture, Design Automation Conference, and International Symposium on Low Power Electronics and Design. Venkatesan received the B.Tech. degree in electronics and communication engineering from the Indian Institute of Technology in 2009 and the Ph.D.degree in electrical and computer engineering from Purdue University in 2014.
\end{IEEEbiography}

\begin{IEEEbiography}
[{\includegraphics[width=1in,height=1.25in,clip,keepaspectratio]{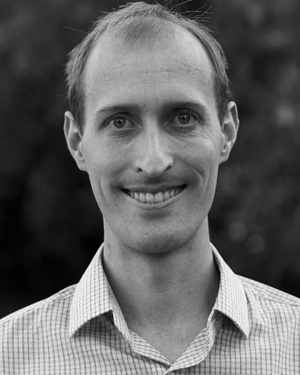}}]{Brian Zimmer}
received the B.S. degree in electrical engineering from the University of California at Davis, Davis, CA, USA, in 2010, and the M.S. and Ph.D. degrees in electrical engineering and computer sciences from the University of California at Berkeley, Berkeley, CA, in 2012 and 2015, respectively. He is currently a Senior Research Scientist with the Circuits Research Group, NVIDIA, Inc., Santa Clara, CA. His research interests include soft error resilience, energy-efficient digital design, low-voltage static random-access memory (SRAM) design, machine learning accelerators, productive design methodologies, and variation tolerance.
\end{IEEEbiography}

\begin{IEEEbiography}
[{\includegraphics[width=1in,height=1.25in,clip,keepaspectratio]{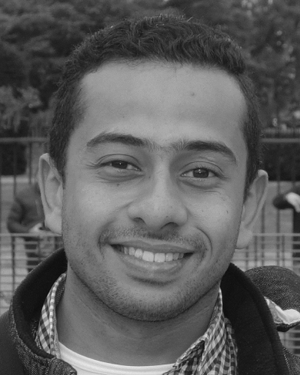}}]{Mustafa Ali}
received the B.Sc. (Hons.) and M.Sc. degrees in electrical engineering from MTC, Cairo, Egypt, in 2011 and 2016, respectively. He is currently pursuing the Ph.D. degree with Purdue University, under the guidance of Prof. Roy. He was honored the Duty Medal for excellent performance during his studies. He worked on flexible electronics applications using TFTs in his M.Sc. degree from 2014 to 2016. In Addition, he worked as a TA and RA at MTC from 2013 to 2017. He was also a Hardware and Embedded Systems Engineer at Integreight, Inc., from 2012 to 2017. He joined the Nano-Electronics Research Lab (NRL), Purdue University, in Spring 2018. His research interests include accelerating brain-inspired computing and machine learning. He is also interested in in-memory computing based on CMOS and post-CMOS devices.
\end{IEEEbiography}

\begin{IEEEbiography}
[{\includegraphics[width=1in,height=1.25in,clip,keepaspectratio]{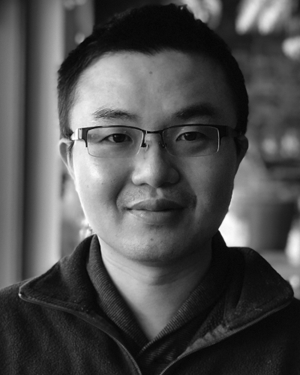}}]{Ming-Yu Liu}
received the PhD degree from the Department of Electrical and Computer Engineering, University of Maryland, College Park, Maryland, in 2012. He is currently a Distinguished Research Scientist and a Manager with NVIDIA Research, Santa Clara, CA, USA. Before joining NVIDIA in 2016, he was a principal research scientist with Mitsubishi Electric Research Labs. His object pose estimation system was awarded one of hundred most innovative technology products by the R\&D magazine in 2014. In CVPR 2018, he won the first place in both the Domain Adaptation for Semantic Segmentation Competition in the WAD challenge and the Optical Flow Competition in the Robust Vision Challenge. His research interests include generative models for image generation and understanding. His goal is to enable machines superhuman-like imagination capabilities.
\end{IEEEbiography}

\begin{IEEEbiography}
[{\includegraphics[width=1in,height=1.25in,clip,keepaspectratio]{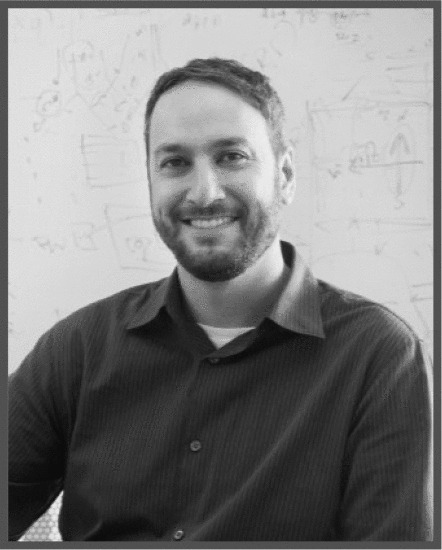}}]{Brucek Khailany}
joined NVIDIA in 2009 and is the Director of the ASIC and VLSI Research group. He leads research into innovative design methodologies for IC development, ML and GPU assisted EDA, and energy efficient ML accelerators. Over 10 years at NVIDIA, he has contributed to many projects in research and product groups spanning computer architecture and VLSI design. Previously, Dr. Khailany was a Co-Founder and Principal Architect at Stream Processors, Inc where he led R\&D related to parallel processor architectures. At Stanford University, he led the VLSI implementation of the Imagine processor, which introduced the concepts of stream processing and partitioned register organizations. He received his PhD in Electrical Engineering from Stanford University and BSE degrees in Electrical and Computer Engineering from the University of Michigan. He is a Senior Member of the IEEE.
\end{IEEEbiography}

\begin{IEEEbiography}
[{\includegraphics[width=1in,height=1.25in,clip,keepaspectratio]{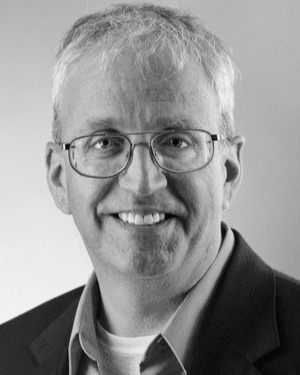}}]{William J. Dally}
is the Chief Scientist and Senior Vice President of research at NVIDIA Corporation, Santa Clara, CA, USA, and an Adjunct Professor and former Chair of computer science at Stanford University, Stanford, CA, USA. His research interests include domain-specific accelerators, parallel computer architectures, interconnection networks, and high-speed signaling circuits. Dally received a Ph.D. degree in computer science in 1986 from the California Institute of Technology, Pasadena, CA, USA. He is a Member of the National Academy of Engineering, and a Fellow of IEEE, ACM, and the American Academy of Arts and Sciences.
\end{IEEEbiography}

\begin{IEEEbiography}
[{\includegraphics[width=1in,height=1.25in,clip,keepaspectratio]{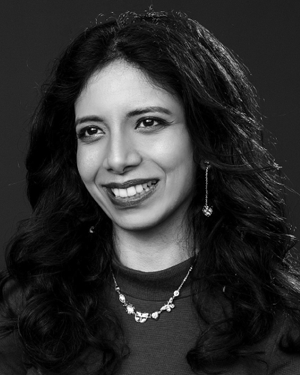}}]{Anima Anandkumar}
is a Bren Professor at Caltech and Senior Director of AI Research at NVIDIA. She is passionate about designing principled AI algorithms and applying them to interdisciplinary domains. She has received several honors such as the IEEE fellowship, Alfred. P. Sloan Fellowship, NSF Career Award, and Faculty Fellowships from Microsoft, Google, Facebook, and Adobe. She is part of the World Economic Forum's Expert Network. Anima received her BTech from Indian Institute of Technology Madras, her PhD from Cornell University, and did her postdoctoral research at MIT and assistant professorship at University of California Irvine.
\end{IEEEbiography}

\clearpage
\newpage

\appendix

\section{Quantization Error Analysis}
\label{apdx:analysis}
Here we present proofs of theorems and lemmas presented in the main paper, as well as some additional details of the empirical evaluations.

\subsection{Proofs}

Before presenting the proofs, we want to clarify the error definition and assumptions introduced in the main paper. Previously, we claim that minimizing $r_{t} = \normsq{\log_2 \abs{W_{t+1}^{U}}  - \log_2 \abs{W_{t+1}}}$ is equivalent to minimizing relative quantization error $\normsq{(W_{t+1} - W_{t+1}^{U}) / W_{t+1}}$. To better understand it, we transform the form of relative quantization error as follows:
\begin{align*}
    &\normsq{(W_{t+1} - W_{t+1}^{U}) / W_{t+1}} \\
    &= \normsq{(I - W_{t+1}^{U}) / W_{t+1}} \\
    &= \normsq{(I - \abs{W_{t+1}^{U})} / \abs{W_{t+1}}} \\ 
    &(sign(W_{t+1}^{U}) = sign(W_{t+1})) \\
    &= \normsq{(I - 2^{\log_2 \abs{W_{t+1}^{U}}  - \log_2 \abs{W_{t+1}}}} \\ 
    &\text{(transfer to base-2 logarithmic space)}
\end{align*}
This relaxation suggests that minimizing $r_{t}$ is equivalent to minimizing the relative quantization error.

We start to introduce the simplified logarithmic quantization we used for the analysis. The stochastic rounding (SR) is defined as follows:
\begin{equation}
\label{eq:stochastic_rounding}
    \SR(x)= \left\{\begin{array}{ll}
\lfloor x \rfloor+1 & \text { for } p \leq x -\lfloor x \rfloor, \\
\lfloor x \rfloor & \text { otherwise,}
\end{array}\right.
\end{equation}
where $p \in [0,1]$ is generated by a uniform random number generator. SR makes sure the rounded number is an unbiased estimate of its full-precision counterpart: $\E \SR(x) = x$, which is an important property for the analysis. 

Equipped with SR, we define the simplified logarithmic quantization function:
\begin{equation}
\label{eq:simplified_logquant}
    \logquant(x) = sign(x) \times 2^{\Tilde{x} / \gamma},
\end{equation}
where $\Tilde{x} = \SR(\, \log_2 \abs{x} \times \gamma)$. We ignore the scale factor and the clamping function to ensure our focus is on the effect of the quantization gap instead of the dynamic range. 

Before proving our main results, we want to introduce an important proposition that describes the error introduced by stochastic rounding.
\begin{restatable}{proposition}{sr_error}\label{thm:sr_error}
For any vector $x$, the quantization error introduced by stochastic rounding $r = \SR(x) - x$ can be bounded in expectation, as:
    \begin{equation}
        \Expect \normsq{r} \leq \sqrt{d} \, \norm{x},
    \end{equation}
    where d is the dimension of $x$.
\end{restatable}
\begin{proof}
    Let $r_i$ denotes the $ith$ element of $r$ and let $q_i = x_i -\lfloor x_i \rfloor$. $r_i$ can be represented as follows:
    \begin{align*}
        r_i &= \left\{\begin{array}{ll}
        \lfloor x_i \rfloor+1 - x_i & \text { for } p \leq x_i -\lfloor x_i \rfloor, \\
        \lfloor x_i \rfloor - x_i & \text { otherwise,}
        \end{array}\right. \\
                    &= \left\{\begin{array}{ll}
        -q_i + 1 & \text { for } p \leq q_i, \\
        -q_i & \text { otherwise.}
        \end{array}\right.
    \end{align*}
    $r_i$ can be bounded by expectation, as:
    \begin{align*}
        \Expect r_i^{2} &\leq (-q_i + 1)^{2} q_i + (-q_i)^{2} (1-q_i) \\
                        &= q_i (1 - q_i) \\
                        &\leq \min\{q_i, 1-q_i\} \\
                        &= \min\{x_i -\lfloor x_i \rfloor, 1-x_i + \lfloor x_i \rfloor\} \\
                        &\leq \abs{x_i}.
    \end{align*}
    Therefore, by summing over index $i$, we can get:
    \begin{align*}
         \Expect \normsq{r} &\leq \norm{x}_1 \\
                            &\leq \sqrt{d} \, \norm{x}.
    \end{align*}
\end{proof}
Now we start to prove Theorem \ref{thm:errorGD} given $U_{GD} = W - \eta \, \nabla_{W}$.
\errorGD*
\begin{proof}
    We know that:
    \begin{multline*}
        \Expect r_{t,GD} = \Vert \log_2 \abs{\logquant(W_{t} - \eta_1 \, \nabla_{W_t})} \\
        - \log_2 \abs{W_{t} - \eta_1 \, \nabla_{W_t}} \Vert^{2}.
    \end{multline*}
    By replacing $\logquant$ with Equation \ref{eq:simplified_logquant}, we can get:
    \begin{multline*}
        \log_2 \abs{\logquant(W_{t} - \eta_1 \, \nabla_{W_t})} = \\
        \frac{1}{\gamma} \SR(\gamma \, \log_2 \abs{W_{t} - \eta_1 \, \nabla_{W_t}}).
    \end{multline*}
    Plug it back to $\Expect r_{t,GD}$, we get: 
    \begin{multline*}
        \Expect r_{t,GD} =\frac{1}{\gamma^{2}} \, \Vert \SR(\gamma \, \log_2 \abs{W_{t} - \eta_1 \, \nabla_{W_t}})  \\
        - \gamma \, \log_2 \abs{W_{t} - \eta_1 \, \nabla_{W_t}} \Vert^{2}.
    \end{multline*}
    Given Proposition \ref{thm:sr_error}, we can upper bound the quantization error introduced by stochastic rounding:
    \begin{align*}
         &\normsq{\SR(\gamma \, \log_2 \abs{W_{t} - \eta_1 \, \nabla_{W_t}})  - \gamma \, \log_2 \abs{W_{t} - \eta_1 \, \nabla_{W_t}}} \\
         &\leq  \sqrt{d} \, \norm{\gamma \, \log_2 \abs{W_{t} - \eta_1 \, \nabla_{W_t}}}.
    \end{align*} 
    Therefore, we can get:
    \begin{align*}
         \Expect r_{t,GD} &\leq \frac{\sqrt{d}}{\gamma^{2}} \, \norm{\gamma \, \log_2 \abs{W_{t} - \eta_1 \, \nabla_{W_t}}} \\
                          &\leq \frac{\sqrt{d}}{\gamma} \, \norm{\log_2 \abs{W_{t} - \eta_1 \, \nabla_{W_t}}}.
    \end{align*}
\end{proof}
Given $U_{MUL} = sign(W) \odot 2^{\Tilde{W} - \eta \, \nabla_{W} \odot sign(W)}$, Theorem \ref{thm:errorMUL} follows a similar proof as Theorem \ref{thm:errorGD}.
\errorMUL*
\begin{proof}
    \begin{multline}
    \label{eq:r_mul_1}
         \Expect r_{t,MUL} = \Vert\log_2 \abs{\logquant(2^{\Tilde{W_{t}} - \eta_2 \, \nabla_{W_{t}} \odot sign(W_{t})})}  \\
         - \log_2 \abs{2^{\Tilde{W_{t}} - \eta_2 \, \nabla_{W_{t}} \odot sign(W_{t})}}\Vert^{2}.
    \end{multline}
    By replacing $\logquant$ with Equation \ref{eq:simplified_logquant}, we can get:
    \begin{multline*}
        \log_2 \abs{\logquant(2^{\Tilde{W_{t}} - \eta_2 \, \nabla_{W_{t}} \odot sign(W_{t})})} \\
        = \frac{1}{\gamma} \, \SR(\gamma \, (\Tilde{W_{t}} - \eta_2 \, \nabla_{W_{t}} \odot sign(W_{t}))).
    \end{multline*}
    Plug it back to Equation \ref{eq:r_mul_1}:
    \begin{multline*}
        \Expect r_{t,MUL} = \frac{1}{\gamma^{2}} \, \Vert\SR(\gamma \, (\Tilde{W_{t}} - \eta_2 \, \nabla_{W_{t}} \odot sign(W_{t}))) \\
        - \gamma \, (\Tilde{W_{t}} - \eta_2 \, \nabla_{W_{t}} \odot sign(W_{t}))\Vert^{2}.
    \end{multline*}
    Because $\Tilde{W_{t}}$ is already an integer, $\SR(\gamma \, \Tilde{W_{t}}) - \gamma \, \Tilde{W_{t}} = 0$, and thus we can eliminate $\Tilde{W_{t}}$ in the equation:
    \begin{multline*}
        \Expect r_{t,MUL} = \frac{1}{\gamma^{2}} \, \Vert\SR(- \gamma \, \eta_2 \, \nabla_{W_{t}} \odot sign(W_{t})) \\
        + \gamma \, \eta_2 \, \nabla_{W_{t}} \odot sign(W_{t})\Vert^{2}.
    \end{multline*}
    Similar to the proof of Theorem \ref{thm:errorGD}, we can upper bound it using Proposition \ref{thm:sr_error}, and get:
    \begin{align*}
        \Expect r_{t,MUL} &\leq \frac{\sqrt{d}}{\gamma^{2}} \, \norm{\gamma \, \eta_2 \, \nabla_{W_{t}} \odot sign(W_{t})} \\
                          &\leq \frac{\sqrt{d} \, \eta_2}{\gamma} \, \norm{\nabla_{W_{t}}}.
    \end{align*} 
\end{proof}

\errorMULsign*
\begin{proof}
    We can simply replace $\nabla_{W_{t}}$ with $sign(\nabla_{W_{t}})$ in the result of Theorem \ref{thm:errorMUL}, and show:
    \begin{align*}
        \frac{\sqrt{d} \, \eta_2}{\gamma} \, \norm{sign(\nabla_{W_{t}})} &\leq \frac{d \, \eta_2}{\gamma}.
    \end{align*}
\end{proof}

\subsection{Evaluations}
As shown in Figure \ref{fig:framework}, we evaluate empirical quantization errors from different learning algorithms when training ResNet-50 on ImageNet. The quantization error is computed at each iteration by $\normsq{\log_2 \abs{W_{t+1}^{U}}  - \log_2 \abs{W_{t+1}}}$. We run each experiment with a full epoch and average the quantization error over iterations. When varying learning rate $\eta$, we fix the base factor $\gamma$ as $2^{10}$. We also fix $\eta$ as $2^{-6}$ when varying $\gamma$.



\section{Multi-Base LNS}
\label{apdx:lns}

\subsection{Conversion Approximation}

We first recap the dot product operation we defined before.
\begin{equation}
\label{eq:dot_product_2}
\begin{split}
    \va^{T} \vb &= \sum_{i=1}^{n} \sign_i \times 2^{\Tilde{\va_i}  / \basefactor} \times 2^{\Tilde{\vb_i}  / \basefactor}, \\
    &= \sum_{i=1}^{n} \sign_i \times 2^{(\Tilde{\va_i}+\Tilde{\vb_i})  / \basefactor}, \\
    &= \sum_{i=1}^{n} \sign_i \times 2^{\Tilde{p_i} / \basefactor},
\end{split}
\end{equation}
where $\sign_i = sign(\va_i) \oplus sign(\vb_i)$.

To understand how we approximate the conversion, we first introduce how ordinary conversion is computed in LNS. Let $\Tilde{p_{iq}}$ and $\Tilde{p_{ir}}$ be positive integers representing quotient and remainder of the intermediate result $\Tilde{p_i}/\basefactor$ in Equation \ref{eq:dot_product_2}, and let $v_r = 2^{\Tilde{p_{ir}}/\basefactor}$. Therefore, 
\begin{equation}
    \begin{split}
        2^{\Tilde{p_i}/ \basefactor} &= 2^{\Tilde{p_i} / \basefactor} = 2^{\Tilde{p_{iq}} + \Tilde{p_{ir}}/\basefactor} = 2^{\Tilde{p_{iq}}} \times 2^{\Tilde{p_{ir}}/\basefactor} \\
        &= (v_r << \Tilde{p_{iq}}),
    \end{split}
\end{equation}
where $<<$ is left bit-shifting. This transformation enables fast conversion by applying efficient bit-shifting over $v_r$ whose value is bounded by the remainder. The different constant values of $v_r = 2^{\Tilde{p_{ir}}/\basefactor}$ can be pre-computed and stored in a hardware look-up table (LUT), where the remainder $\Tilde{p_{ir}}$ is used to select the constant for $v_r$. The quotient $\Tilde{p_{iq}}$ then determines how far to shift the constant. Furthermore, because $\basefactor = 2^b$, the least significant bits (LSB) of the exponent are the remainder and the most significant bits (MSB) are the quotient.
As the size of the LUT grows, the computational overhead from conversion increases significantly. Typically, the LUT is required to contain $2^b$ entries for storing all possible values of $v_r$, which can be a large overhead for large values of $b$. 

A straightforward solution for reducing the size of LUT is utilizing Mitchell approximation~\cite{5219391}: $v_r = 2^{\Tilde{p_{ir}}/2^b} = (1 + \Tilde{p_{ir}}/2^b)$. However, if $v_r$ is far away from zero or one, the approximation error induced by Mitchell approximation will be significant. To alleviate this error, we propose a hybrid approximation that trades off efficiency and approximation error. Specifically, we split $p_{ir}$ into $p_{irM}$ and $p_{irL}$ to represent the MSB and LSB of the remainder, respectively. LSB values $2^{\Tilde{p_{irL}}}$ are approximated using Mitchell approximation, and MSB values $2^{\Tilde{p_{irM}}}$ are pre-computed and stored using LUT, such that:
\begin{equation}
\label{eq:lns_approx}
    \begin{split}
     v_r = 2^{\Tilde{p_{ir}}/2^b} &= 2^{\Tilde{p_{irM}}/2^b} \times 2^{\Tilde{p_{irL}}/2^b} \\
     &= (1 + \Tilde{p_{irL}}/2^b) \times 2^{\Tilde{p_{irM}}/2^b},
    \end{split}
\end{equation}
where $p_{irM}$ and $p_{irL}$ represent $b_m$ MSB and $b_l$ LSB bits of $p_{ir}$. This reduces the size of LUT to $2^{b_m}$ entries. For efficient hardware implementation, we use $2^{b_m}$ registers to accumulate different partial sum values and then multiply with constants from the LUT. 

\subsection{Approximation-Aware Training}
\begin{table}[h!]
  \centering
  \caption{Benchmarking conversion approximation for multi-base LNS}
  \label{tab:standard_approxi}
  \vskip 0.1in
  \tabcolsep=0.1cm
  \small
\begin{tabular}{lcccc}
    \toprule
     &
    \textbf{LUT=1} &
    \textbf{LUT=2} &
    \textbf{LUT=4} &
    \textbf{LUT=8} \\
    \midrule
    \\[-1em]
    \textbf{CIFAR-10} & \multirow{2}{*}{92.58} & \multirow{2}{*}{92.54} & \multirow{2}{*}{92.68} & \multirow{2}{*}{\textbf{93.43}} \\
    \small{(Accuracy)} & & & & \\
    \\[-1em]
    \hline
    \\[-1em]
    \textbf{ImageNet} & \multirow{2}{*}{75.80} & \multirow{2}{*}{75.85} & \multirow{2}{*}{75.94} & \multirow{2}{*}{\textbf{76.05}} \\
    \small{(Accuracy)} & & & & \\
    \hline
    \\[-1em]
    \textbf{SQuAD} & \multirow{2}{*}{87.57} & \multirow{2}{*}{87.11} & \multirow{2}{*}{\textbf{88.00}} & \multirow{2}{*}{87.82} \\
    \small{(F-1)} & & & & \\
    \hline
    \\[-1em]
    \textbf{GLUE} & \multirow{2}{*}{84.89} & \multirow{2}{*}{85.48} & \multirow{2}{*}{86.93} & \multirow{2}{*}{\textbf{88.15}} \\
    \small{(F-1)} & & & & \\
    \hline
    \\[-1em]
    \textbf{Energy Cost} & \multirow{2}{*}{12.29} & \multirow{2}{*}{14.71} & \multirow{2}{*}{17.24} & \multirow{2}{*}{19.02} \\
    \small{(fJ / op)} & & & & \\
    \bottomrule
    
  \end{tabular}
\end{table}
We demonstrate that the approximation error induced by the conversion approximation does not affect the overall accuracy significantly. The deterministic approximators can be viewed as additional non-linear layers in networks, and thus they can be learned during training. Because this process is similar to quantization-aware training, we denote it as approximation-aware training. To verify our hypothesis, 
we simulate the proposed conversion approximation in LNS. The approximators are only applied to the forward propagation to allow approximation-aware training. After training, an approximated model can also be deployed for fast inference. With 8-bit and base factor $\gamma=8$, we evaluate the approximation setting from $LUT=1$ to $LUT=8$. We use the BERT-base model for SQuAD and GLUE in this experiment, and we also report the energy cost per operation for each approximation setting.  As shown in Table \ref{tab:standard_approxi}, the approximated networks achieve almost no loss of accuracy while reducing energy cost by 35\% maximally.  

\section{Experiments}





\subsection{Datasets and Models}

\subsubsection{ResNet Models} 
We use residual networks for benchmarks on image datasets \cite{ResNet}. We quantize all fully-connected and convolutional layers in ResNet, including both forward and backward propagation. Besides, we leave batch-norm layers at full-precision for simplicity. SGD optimizer is applied by default with its standard learning rate schedule.

\subsubsection{BERT Models} 
We perform quantization on pre-trained BERT models for language fine-tuning tasks. BERT models are the state-of-the-art language representation models which include 110M parameters in the BERT-base model and 320M parameters in the BERT-large model \cite{devlin2018bert}. We quantize all GEMM operations for both models, which consist of 99\% of all parameters. AdamW optimizer is applied by default.

\subsubsection{CIFAR-10}
We use ResNet-18 to evaluate different quantization settings on CIFAR-10 dataset \cite{cifar}. CIFAR-10 consists of 60,000 images in 10 different classes. The network is trained for 300 epochs, and we use a fixed learning rate decay schedule that decayed every 100 epochs. Besides, we use a tuned SGD optimizer by default, where the initial learning rate is 0.1,  weight decay is 0.0001, and momentum is 0.9.

\subsubsection{ImageNet} 
The ILSVRC2012 ImageNet dataset consists of 1.2 million images belonging to 1,000 classes \cite{deng2009imagenet}. We use Resnet-50 as the base model, and the network is trained for 90 epochs for all settings. Similarly, we use a tuned SGD optimizer with a learning rate warmup by default, where the configuration is the same as the one in the CIFAR-10 experiment. For LNS-Madam optimizer, as mentioned in previous studies, multiplicative learning algorithms may enjoy a weight initialization different from the standard practice. Therefore, on the ImageNet benchmark, we use SGD as a warm-up for the first 10 epochs to mitigate this initialization effect.

\subsubsection{SQuAD}
The Stanford Question Answering Dataset (SQuAD v1.1) is a collection of 100k crowdsourced question/answer pairs \cite{squad}. We evaluate our framework on SQuAD and use the generic pre-trained BERT and BERT-large models for the fine-tuning (\cite{devlin2018bert}). The maximum sequence length and document stride are also set to be 384 and 128, respectively. We fine-tune the network for 2 epochs and use a tuned AdamW optimizer by default, where its learning rate starts from 0.00003.

\subsubsection{GLUE}
The General Language Understanding Evaluation (GLUE) benchmark is a collection of diverse natural language understanding tasks \cite{wang-etal-2018-glue}. We use a pre-processing setting similar to the setting for SQuAD and fine-tunes using both BERT-base and BERT-large models for 2 epochs. AdamW optimizer is applied by default, where its initial learning rate is 0.00002.

\end{document}